\documentclass[journal]{IEEEtran}

\usepackage[switch]{lineno}
\usepackage{ieee_thm} 
\usepackage{des} 
\usepackage[brazilian,english]{babel}
\usepackage[usenames]{color}
\usepackage[cmex10]{amsmath}
\usepackage{color,graphicx,psfrag}
\usepackage{amssymb,stmaryrd,soul,amsmath}
\usepackage{comment}
\usepackage[noend]{algpseudocode}
\usepackage{subfigure}
\usepackage[dvipsnames]{xcolor}
\usepackage{float}
\usepackage{array}
\usepackage{caption}
\usepackage{multirow}
\usepackage[ruled,vlined,linesnumbered]{algorithm2e}

\def\cor{black}
\def\painter#1{\textcolor{\cor}{ #1}}
\def\painterX#1{\color{\cor}{ #1}}

\ifCLASSINFOpdf
\else
\fi

\makeindex

\begin{document}


\title{
Estimating action plans for smart poultry houses
}

\author{
	Darlan Felipe~Klotz$^{1}$ \quad 
	Richardson Ribeiro$^{1}$ \quad 
	Fabrício Enembreck $^{2}$ \quad
	Gustavo Denardin $^{1}$ \quad
	Marco Barbosa $^{1}$ \quad
	Dalcimar Casanova$^{1}$\quad
	Marcelo Teixeira$^{1}$
	\thanks{This work has been supported 
		by the \emph{National Council for Scientific and Technological Development} (CNPq), under grant number 402145/2016-0, by CAPES, FINEP and Araucária Foundation.}
	\thanks{$^{1}$The authors are with the Graduate Program in Electrical Engineering of the Federal University of Technology Paraná, Pato Branco, Brazil
		({\tt\small darlanklotz@gmail.com), (\{richardsonr \bcom gustavo \bcom mbarbosa \bcom dalcimar\bcom marceloteixeira\}@utfpr.edu.br}).
		}
	\thanks{$^{2}$The author is with the Graduate Program in Informatics of the Pontifical Catholic University of Paraná, Curitiba - PR, Brazil
		({\tt\small fabricio@ppgia.pucpr.br).
		}
}}

\maketitle

\begin{abstract}
In poultry farming, the systematic choice, update, and implementation of periodic ($t$) action plans define the feed conversion rate (\FCR[t]), \painter{which is an acceptable measure} for successful production. 
\painter{Appropriate action plans provide} tailored resources for broilers, allowing them to grow within the so-called thermal comfort zone, without wast or lack of resources. 
Although the implementation of an action plan is automatic, its configuration depends on the knowledge of the \painter{specialist, tending to be inefficient and error-prone, besides to result in different \FCR[t] for each poultry house}.
In this article, we claim that the \painter{specialist}'s perception can be reproduced, to some extent, by computational intelligence. 
\painter{By combining deep learning and genetic algorithm techniques, we show how action plans can adapt their performance over the time, based on previous well succeeded plans. 
We also implement} a distributed network infrastructure that allows to replicate our method over distributed poultry houses, for their smart, interconnected, and adaptive control. A supervision system is provided as interface to users. \painter{Experiments conducted over real data show that our method improves 5\% on the performance of the most productive specialist, staying very close to the optimal \FCR[t]}.
\end{abstract}


\IEEEpeerreviewmaketitle

\section{Introduction}\label{sec:introd}

\emph{Poultry farming} is the agro-industrial sector focused on producing chicken meat for consumption. 
Worldwide, the activity has attracted attention, and investments, due to its economic and social role, aligned with the continuously increasing world demand for chicken meat. 
According to the last published survey, only in 2019 the USA was responsible for producing 19.941 million tons of chicken meat, followed by Brazil and China with 13.245 and 13.750 million tons, respectively \cite{ABPA}. 

Technically, the birds raise in infrastructures called \emph{poultry houses} (\houses), which include a set of variables that are automatically controlled by electronic devices. 
\painter{The daily recipe prescribing the numeric parameters for the automatic controller is called \emph{action plan} or \emph{poultry management} \cite{gleaves1989application,book:broiler}. 
The quality of an action plan reflects the general welfare for birds, allowing them to grow up under appropriate conditions of feeding, housing, and health. In this paper, in particular, an action plan is exploited in terms of its environment-related variables. 
We seek for combined indoor climate conditions that allow chickens to grow up under \emph{thermal comfort}, achieved within the so-called \emph{comfort zone} \cite{botreau2007aggregation}. 
The comfort zone directly affects productivity, which is acceptably measured by periodic ($t \in \mathbb{N}$) \emph{feed conversion rates} (\FCR[t]) estimations \cite{skinner2004components}.
The more accurate the action plan, the lower the accumulated \FCR[t], and the greater the production.
}

\painter{
Recent advances on improving welfare, and consequently the \FCR[t], have tackled 
health and behaviour \cite{xiao2019behavior,Henriksen}, 
nutrition \cite{chen2016impact,trocino2015effect},
genetics \cite{hoffmann2005research}, and 
thermal comfort \cite{Ribeiro_action_plans,lorencena2019framework}, etc. 
However, a remaining barrier in broiler production is the lack of tools to support the specialist with the systematic choice, update, and implementation of daily action plans.
}

Although the practical implementation of an action plan is automatic, the configuration of its variables depends essentially on the knowledge accumulated by the \painter{specialist}. Therefore, this decision is empiric and tends to be complex, inefficient, and error-prone \cite{Ribeiro_action_plans}. 
This is even more perceptible when a \house is part of a larger system of \houses which, although appears as a promising model for broilers production, also leads different specialists to conflict their perceptions. As they are subject to different management situations, they tend to build their own personalised action plans. 

\painter{
These unsystematic interventions on the process may return a distinct \FCR[t] for each \house, which makes unclear the overall performance estimation. 
In contrast, even minor variations in \FCR[n] (at the end of a n-days flock production) may cause considerable losses. 
Suppose, for example, that after 40 days the \FCR[40] is 0.1 greater than the expected. 
This implies that each bird just ate 0.1 \Kg (or 100 \grama) more feed to reach 1 \Kg of weight. 
Then, a \house comprising about 35 thousand broilers will consume 3.5 tons more feed than expected. 
}

From this perspective, the literature presents some initiatives that apply computational methods to derive intelligent action plans. For example, \cite{Ribeiro_action_plans} proposes a bio-inspired model based on \emph{Artificial Neural Networks} (\ANNs) to suggest action plans for the specialist based on the climatic conditions of the \house. However, this approach is not integrated with the process controller, so that it cannot reconfigure its actions dynamically according to the \ANN suggestion. Furthermore, the approach in \cite{Ribeiro_action_plans} does not consider the dynamic adaptation of its estimations over the time. 

In this article, the specialist perception about action plans configurations is captured and reproduced, to some extent, by computational intelligence. This is similar to \cite{Ribeiro_action_plans} but we use a more suitable memory-aware model based on \emph{Long short-term memory} (\LSTM) networks, instead of \ANNs. 
Furthermore, we show how estimated action plans can be combined across algorithms that artificially emulate genetic mutations. 
This results in improved action plans that are capable of adapting themselves over the time based on previously successful action plans. 
By replicating our method over a set of \houses, we solidify a new foundation to support smart, cooperative, distributed, and interconnected poultry farming. 

\painter{
Experiments are conducted over real data in order to illustrate our contributions. 
Results indicate that our method estimates a $\FCR[40]$ of $1.5610$. 
In comparison, the $\FCR[40]$ resulting from the progressive application of empiric action plans, chosen by the specialist who had the best performance among all other specialists in the evaluated group, under the same setup, was of $1.640$, therefore $0.079$ (about $5\%$) worse than the model suggestion. Furthermore, our method shows proximity from the best performance achievable by Ross-breed broilers, which is $\FCR[40] = 1.558$ \cite{rossperformace:Online}. 
}



In the following, 
\Sect~\ref{sec:back} reports the background on poultry farming, practices, systems, and architectures;
\Sect~\ref{sec:intel} presents the intelligent methods used in the paper;
\Sect~\ref{sec:main} introduces our main results, which are implemented in \Sect~\ref{sec:implement}; and 
\Sect~\ref{sec:conc} discusses some conclusions.

\section{Poultry Farming Foundations} \label{sec:back}

This section presents general aspects of poultry farming, including physical layouts of production, control systems options, and the formal structuring of action plans that guide the daily practices of production management. 

\subsection{Poultry farming architectures}

\begin{figure}[!htb]
	\psfrag{Input}[c][c]{\footnotesize{Input}}
	\psfrag{Output}[c][c]{\footnotesize{Output}}
    \psfrag{Ar}[c][c]{\footnotesize{Air}} 
	\psfrag{Sensors}[c][c]{\footnotesize{Sensors}} 
	\psfrag{Ar_quente}[c][c]{\footnotesize{Hot air}} 
	\psfrag{Temperture}[c][c]{\footnotesize{Temperature}}
	\psfrag{Humidity}[c][c]{\footnotesize{Humidity}} 
	\psfrag{Fornalhas}[c][c]{\footnotesize{Furnaces}} 
	\psfrag{Exaustores}[c][c]{\footnotesize{Exhausting Fans}}
	\psfrag{Courtains}[c][c]{\footnotesize{Curtains}} 
	\centerline{
		\includegraphics[width=7.5cm]{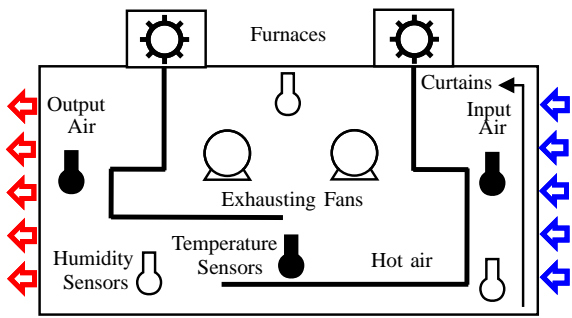}
	}
	\caption{Architecture for broiler housing.}
	\label{fig:house}
\end{figure}

\painter{
The controlled environment in which birds are raised is shaped as a rectangular (usually 16\meter wide, 150\meter long, with 4\meter high sidewalls) fully encapsulated structure called a \emph{poultry house} (\house). 
Internally, a \house includes electronic devices such as sensors, actuators, and control technology. 
\Fig~\ref{fig:house} illustrates a \house architecture known as \emph{tunnel ventilation}. 
Other layouts can be found in \cite{poultry:Online}. 
The results to be presented in this paper are independent of architecture. 
}

\painter{
Generally, the external air enters at one side and leaves the \house by exit points crafted at the opposite side. Automatic curtains control the air intake, while a coordinated set of exhausting fans controls its outtake, which serves both for cooling and removal of gases. Humidifiers, such as nebulizers and pad-coolings, synchronised with the airflow, spreads water either for cooling or humidity control. Under low temperatures, combustion-based furnaces are used to heat the \house by conducting hot air through valved pipes \cite{brewer1981heating,zhu2005co}.
}

\subsection{Automatic control of poultry houses}

{\painterX
The automatic control system is responsible for the orchestration of exhausters, humidifiers, heaters, and curtains. Together, they allow thermal comfort. 
The controller takes as input the information collected from sensors, implements an brain-like actuation logic, and generates an output that corresponds to electrical signals that activate physical actuators to correct the climate condition as intended. 

Two major concerns challenge the automatic control and automation of poultry houses: 
(i) the synthesis of a control logic; and 
(ii) the progressive setup of the parameters for the controller. 
Parameters, therefore, can be seen as the values to be assigned to control each variable of an action plan. 
}

The complexity behind (i) emerges from the fact that multiple interdependent events must be considered concurrently, in order for an output control action to be synthesised. The use of formal methods can alleviate this burden \cite{lorencena2019framework} and results in a controller that can be implemented by hardware, such as a \emph{PLC} or \emph{microcontroller} \cite{carey1995poultry,glatz2013poultry,fairchild2005basic}.

Concern (ii) appears when a controller technically is ready for use, but it still requires to be set up with parameters that quantify its actuation. Each possible set of parameters represents a potential action plan to be applied during a short period of time, usually a day. Estimating the action plan is, therefore, a combinatorial, multi-variable problem whose solution depends essentially on the \painter{specialists. Furthermore, action plans differ at each step of broilers production, and they must be updated properly.} 

\subsection{Action plans formalisation}

{\painterX
It is usually challenging to discover appropriate matches of climate variables that result in thermal comfort. Particularly, in regions with mixed weather, these variables are subject to extreme variations and can change very quickly, making thermal comfort complex to be derived under unclear, heterogeneous, and unstable conditions \cite{vieira2011preslaughter,glatz2013poultry}. 
Not rarely, this leads specialists to overlap the automatic controller in the hope that a manual intervention could be a short-cut to thermal stabilisation. In general, this is rather inefficient and error-prone, besides to be uncomfortable as it requires the specialist to be full time connected to the process, to external conditions, and to the broilers behaviour. 

In contrast, choosing appropriate action plans is essential in poultry farming. 
Among the variables that form an action plan, temperature and humidity are shown to be the most important ones to be controlled, as they influence directly on productivity \cite{may2000effect}. In fact, broilers metabolism is extremely sensitive to climate variations \cite{Prince,donkoh1989ambient}. While high variations are associated with stress, balanced conditions suggest efficiency \cite{balnave1998increased}.
}



A usual measure to quantify efficiency in broilers production is based on periodic ($t$) samples of \emph{Feed Conversion Rate} (\FCR[t]), formalised as:
\begin{equation} \label{eq1}
\FCR[t]=\frac{\dFC[t]}{\NlB[t] \cdot \MdW[t]}\, \bcom  
\end{equation}
where 
\dia is an integer associated with the period of observation (in days); 
\dFC[t] \painter{is the amount of \emph{feed consumption} until day \dia}; 
\NlB[t] \painter{is the number of \emph{living birds}} at day \dia; and 
\MdW[t] is the \emph{mean weigh} at day \dia \cite{skinner2004components}. 

An effective action plan is achieved when the \FCR[t] is improved, i.e., when it approximates to zero \cite{fontana1992effect}. 
The more accurate the action plan, the lower the conversion rate, and the greater the production.

However, obtaining a satisfactory \FCR[t], at the end of a flock production (usually $\dia = 40$), requires enormous ability from the \painter{specialist} to conduct the daily management. It depends, basically, on how the daily parameters are informed to the controller so that, in combination, they could result in a reasonable accumulated \FCR[40] for the entire flock. 
This task is supported, to some extent, by general guides provided by companies that centralise production.

\painter{
In practice, however, the guides are far away from the real needs faced by the \painter{specialist} during the daily management. 
A general guide does not differentiate, for example, seasonality or geographical peculiarities that cause quick variations of temperature, wind direction, altitude, humidity, and pressure. In addition, each region may have a local legislation (such as environmental issues or health surveillance) which may require customised management practices.
}




As a consequence, \painter{specialists} use their own experience, from previous flocks, to set up action plans. 
This leads to fully (and foolish) empirical decisions, such as the one observed in periodical in-loco visits, that contrast radically to the protocol of best practices \cite{Hi_Pro}: 
(i) induce to scarce or excessive \painter{feed}; 
(ii) \painter{alter cold or heat unnecessarily}; and 
(iii) manage humidity levels according to human sensation.


In fact, in addition to his technical knowledge, patience, empathy, and dedication, the \painter{specialist} also uses his 5 senses to constantly check the birds welfare, monitoring whether they are eating, playing, drinking, chirping, resting, or huddling \cite{book:broiler}. 
By hearing, he checks breathing sound, vocalisation, sneaking, and sneezing, as well as whether or not the mechanical noise produced by the power and ventilation systems are appropriate. 
By nosing, the \painter{specialist} checks the ammonia levels, the \painter{feed} smell, and the air quality. 
By tasting, \painter{he checks feed and water}. 
By touching, he can check the morphology, such as breast conformance and feather condition; he can also feel the level of air circulation on his skin, and the temperature inside and outside the \house; he can also checks the texture of the \painter{feed} (if the pellets break down easily in the hand and in the feeder), and the condition of the litter to possibly identify excessive humidity that could indicate lack of ventilation \cite{fontana2017sound,mollah2010digital,louton2018animal}.

Observe, therefore, that \painter{in addition to guides support there is a number of decisions that do still depend on human perception, experience, education, training, and other technical skills}. 
As a result, performance is quite unpredictable and incomparable in poultry farming. 
This is even more evident when a \house is part of larger production system, as follows. 

\subsection{Distributed poultry houses}

{\painterX
In practice, it is quite usual for a specialist to handle more than one \house into a shared infrastructure. 
This allows for large-scale production, simplifies supply chain, facilitates maintenance, reuses sanitary guards, etc., composing a business model known as \emph{vertically integrated production} \VIP \cite{henry1960integration}.

Despite promising advantages, the potential of \VIP models is still not fully exploited in the literature. 
Currently, each \house is seen as a local, self-contained, entity that does not depend or share any of its features with others. This prevents them to cooperate in order to maximise production or improve practices over the time. 
We claim that the main barrier is technological, which composes the main subject of this paper.}

From now forward, whenever two or more \houses share the same physical infrastructure, we call them a \emph{ poultry condominium} (\PC).
In \PCs, the concept of \emph{connectivity} plays an essential role. 
Nevertheless, only a few initiatives have been reported in the literature to assist poultry production with data exchange. 
In \cite{ammad2014wireless,choukidar2017smart,lashari2018iot}, connectivity is exploited to monitor climate variables in houses, such as temperature and humidity, using wireless sensors. 
However, the connectivity-based cooperation among different \houses, in an attempt for optimising their action plans, has not yet been reported. 

In contrast, some methods are proposed to assist the \painter{specialists} with candidate action plans that can be used for optimisation purposes. However, they are not based on interconnection of \houses. 
For example, \cite{mirzaee2015comparison} develops a climate monitoring and management system for \houses using fuzzy logic. 
This approach performs well \painter{in comparison with other classical controllers, but it does not directly apply to \PCs. 
In \cite{Ribeiro_action_plans} a novel \ANN-based approach in introduced to suggest optimised action plans to the \painter{specialist}. However, it is not integrated with automatic controllers, which prevents control reconfiguration directly from the estimated action plans. Furthermore, this contribution does not propose an adaptive method.}

Therefore, we argue that the existing approaches in the literature are not extensible enough to cover \PC-based architectures for poultry farming. A question that remains is how to estimate, for a set of different and distributed \houses, a reasonable action plan that could be replicated automatically to each controller within the \PC? 
Could this action plan adapt autonomously over time to different environment profiles? 
In the following, we present the foundations that help us to answer these questions.

\section{Artificial Neural Networks and Genetic Algorithms Foundations}\label{sec:intel}

In this section, we provide a general overview on neural networks and genetic algorithms, evidencing their combined role for results to be derived later in \Sect~\ref{sec:main}.

\subsection{Artificial Neural Networks}

\emph{Artificial Neural Networks} (\ANNs) are computational resources that use the notion of biological neurons to emulate the human brain. 
By multiplying and iterating through artificial neurons, a \ANN can anticipate, with reasonable accuracy, the human reasoning to take decisions, so that problems can be solved in advance \cite{wasserman1989neural}. 
An \ANN topology that is used in this article is detailed in the following. 

\subsection{Recurrent Neural Networks}
The main limitation of traditional \ANNs is that they are not forged to accumulate memory, i.e., information about previous events. 
The so called \emph{Recurring Neural Networks} (\RNNs) extends the ordinary \ANNs with this feature. 
\RNNs are alternatives that allow recognizing patterns in data streams, by taking \emph{time} and \emph{order} into account \cite{goodfellow2016deep}.
They are looped networks that allow information to persist, which better imitates the human brain in which knowledge is accumulative \cite{goodfellow2016deep}. 
\Fig~\ref{fig:compRNNandANN} compares the neuron model of a \RNN (\Fig~\ref{fig:rnn}) with a \emph{Feed-Forward}-based \ANN (\Fig~\ref{fig:ann}).

 \def\scaleIMG{0.32}
\begin{figure}[!htb]
	\psfrag{in}{\scriptsize{Input}}
	\psfrag{out}{\scriptsize{Output}}  
	\psfrag{rnn}{}
	\psfrag{ann}{}
    \centering
	    \subfigure[][{\RNN}]{
		\includegraphics[scale = \scaleIMG]{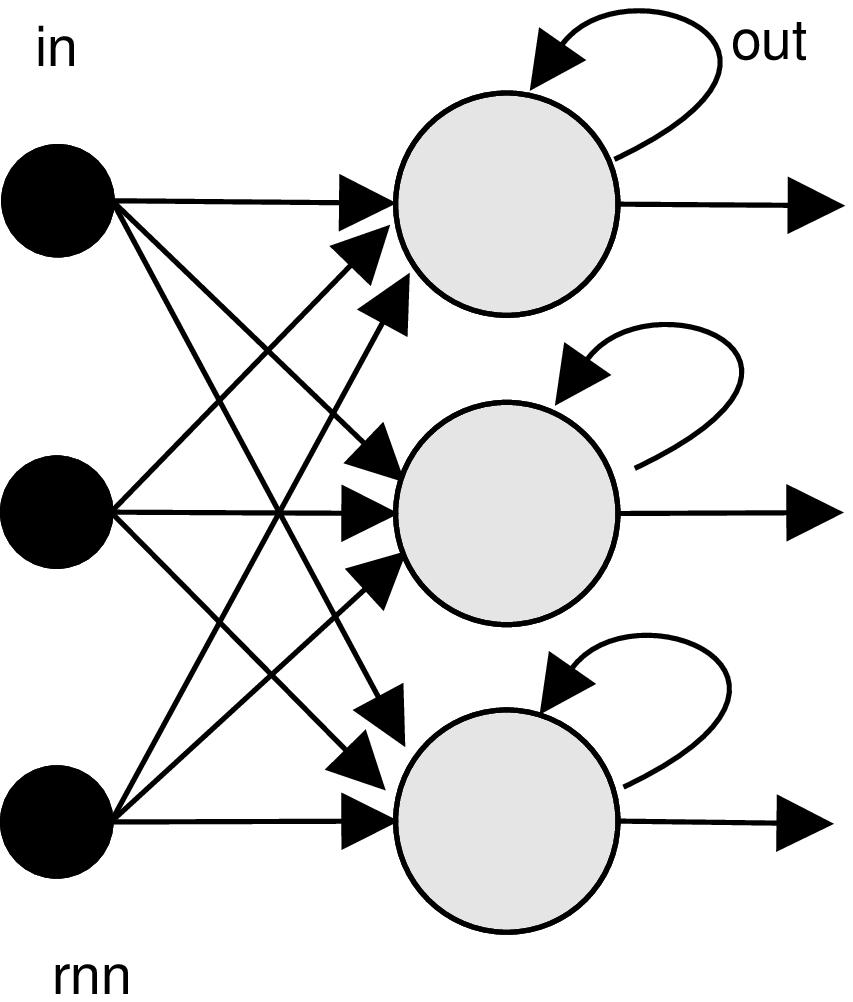}
		\label{fig:rnn}
	}\qquad
		\subfigure[][{\emph{Feed-Forward} \ANN}]{
		\includegraphics[scale = \scaleIMG]{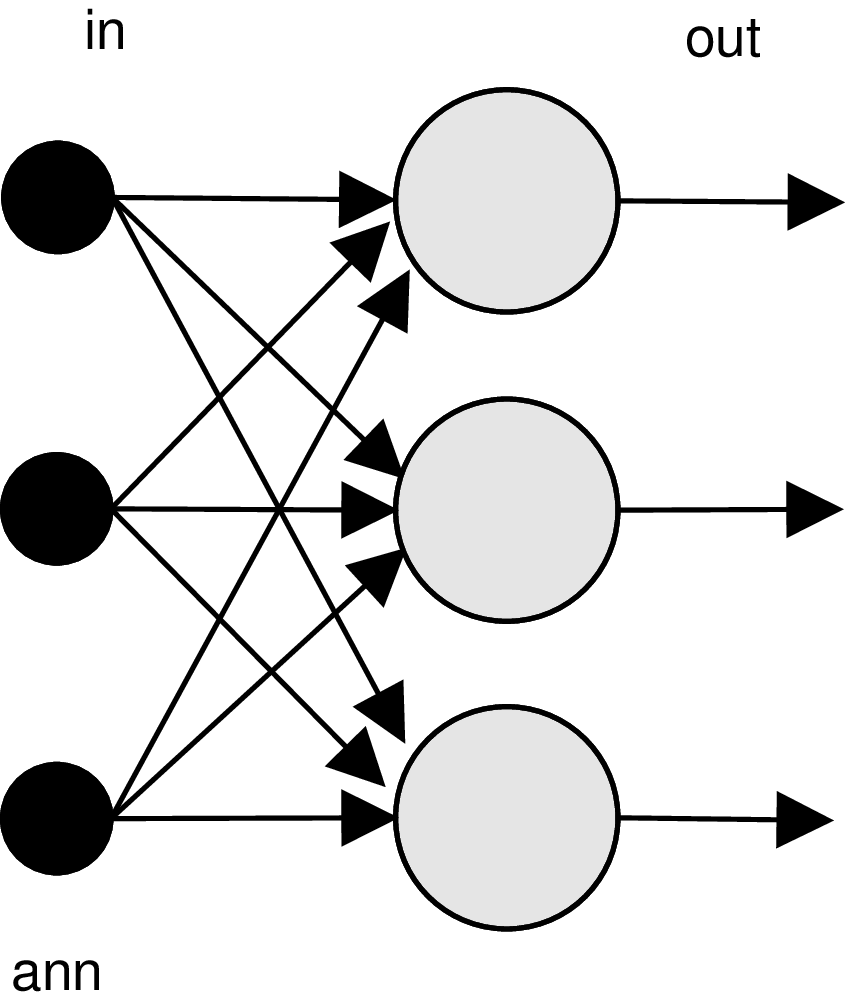}
		\label{fig:ann}
	}
\caption{Architecture of \RNNs and \ANNs.}
\label{fig:compRNNandANN}
\end{figure}
 
The \RNN loops imply that decisions reached at time step $t-1$ affect the decision reached a moment later, at time step $t$. Thus, \RNNs have two input sources: the present and the recent past, which are combined to determine how they respond to new data. 
This memory information is kept within the hidden states of a \RNN, which can pass through many stages of time, as it progresses sequentially. 
The relation between events is called \emph{long-term} dependency, because an event in time is a function of one or more past events \cite{deeplearning:Online}. 
Formally, the process of taking the memory forward is displayed as:
\begin{equation}
 \label{eq2}
\out[t]=\Sig[h]{(\Mat[W]_{xh} \inp[t]+\Mat[W]_{hh} \pout[t] + \Vet[b]_{h})\, ,}
\end{equation}
where 
$\out[t]$ is the hidden state vector in time step $t$; 
$\inp[t]$ is the input data vector in time step $t$; 
$\pout[t]$ is the vector hidden state of the previous step;
$\Mat[W]_{xh}$ is the matrix of the network weights to the input $\inp[t]$;
$\Mat[W]_{hh}$ is the matrix of the network weights to the hidden input $\pout[t]$; 
$\Vet[b]_{h}$ is the vector \emph{bias}; and
$\Sig[h]$ is the activation function \cite{deeplearning:Online}.

The learning of the weight matrices $\Mat[W]_{xh}$ and $\Mat[W]_{hh}$ can be given by the \emph{backpropation through time} method or similar \cite{werbos1990backpropagation}.
However, ordinary \RNNs are unable to capture information from \emph{long-term} dependencies \cite{goodfellow2016deep}, which can be alternatively addressed as follows.

\subsection{Long Short Term Memory} \label{subsec:LSTM_review}

\emph{Long Short Term Memory} (\LSTM) networks are derived from ordinary \RNNs and explicitly designed to avoid the long-term dependency problem, i.e., they are forged to remember information for long periods \cite{hochreiter1997long}. 

\begin{figure}[!htb]
	\psfrag{input}[c][c]{\scriptsize{Input}}
    \psfrag{output}[c][c]{\scriptsize{Output}} 
	\psfrag{gate}[c][c]{\scriptsize{Gate}} 
	\psfrag{State_cell}[c][c]{\scriptsize{Cell State}}  
	\psfrag{forget}[c][c]{\scriptsize{Forget}} 
	\psfrag{p1}[c][c]{\scriptsize{$\Vet[i]^{\left \langle t \right \rangle}$}}
	\psfrag{p2}[c][c]{\scriptsize{\Hip}} 
	\psfrag{s1}[c][c]{\scriptsize{$\Vet[f]^{\left \langle t \right \rangle}$}}
	\psfrag{s2}[c][c]{\scriptsize{$\Vet[j]^{\left \langle t \right \rangle}$}}
	\psfrag{s3}[c][c]{\scriptsize{$\Vet[o]^{\left \langle t \right \rangle}$}}
	\psfrag{xt}[c][c]{\scriptsize{$\inp[t]$}}  
	\psfrag{ht1}[c][c]{\scriptsize{$\pout[t]$}}
	\psfrag{ht}{\scriptsize{$\out[t]$}} 
    \psfrag{c1}[c][c]{\scriptsize{$\textbf{C}^{\left \langle t-1 \right \rangle}$}}
	\psfrag{c}{\scriptsize{$\textbf{C}^{\left \langle t \right \rangle}$}} 
	\centerline{
		\includegraphics[scale=0.5]{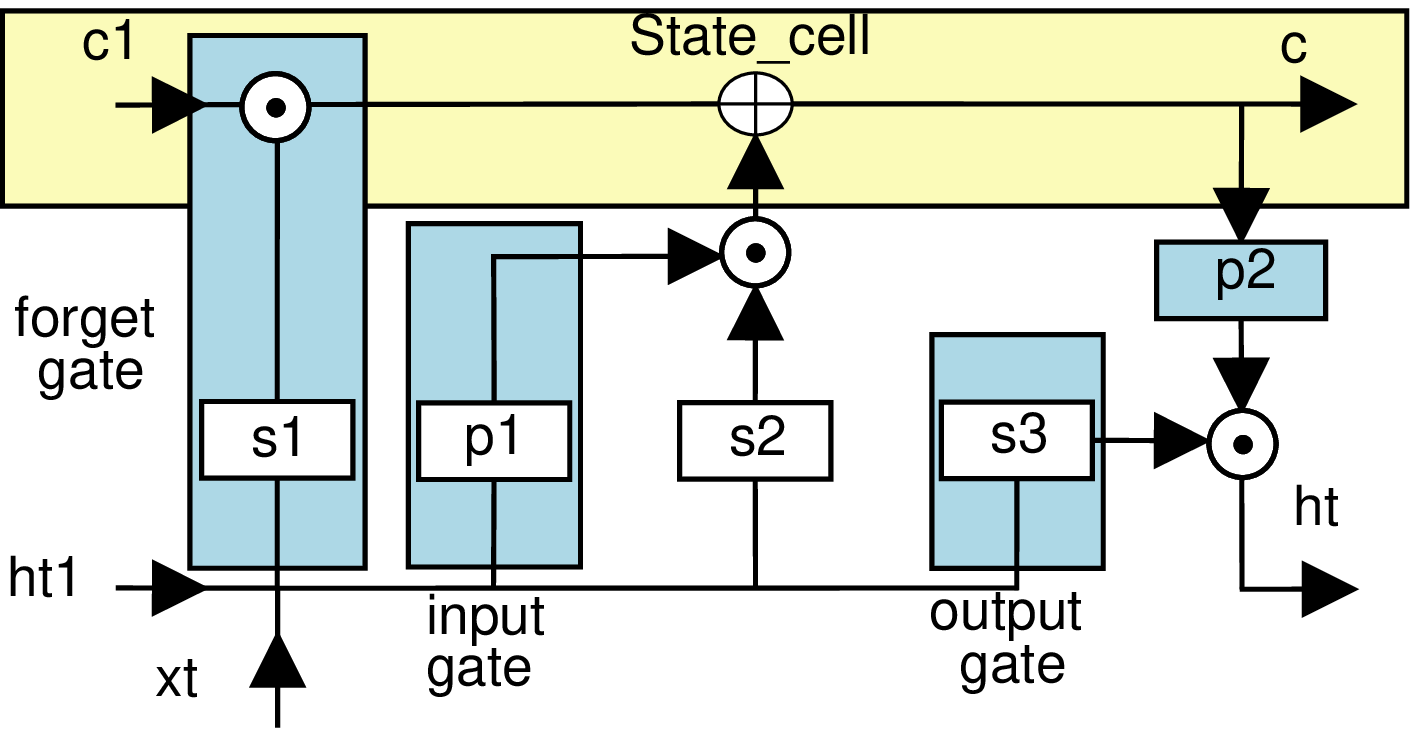}
	}
	\caption{\LSTM network architecture.}
	\label{fig:lstm}
\end{figure}

\painter{
A \LSTM is composed of: 
(i) a \emph{hidden} state, 
(ii) a \emph{cell} state, and 
(iii) access \emph{gates}. 
The hidden state in the current step ($\out[t]$) contains the output for that step. 
The cell state contains information learned from the previous steps. 
At each step, information is added to, or removed from, the cell state using three gates, i.e., gates select and save relevant information, eliminating the others. \Fig~\ref{fig:lstm} illustrates a \LSTM network.}

Formally, the architecture can be exposed as follows:
\begin{equation}
 \label{eqLSTM_1}
\Vet[f]^{\left \langle t \right \rangle}=\Sig[f](\Mat[W]_{xf} \inp[t] + \Mat[W]_{hf} \pout[t] + \Vet[b]_f)\, ;
\end{equation}
\begin{equation}
 \label{eqLSTM_2}
\Vet[i]^{\left \langle t \right \rangle}=\Hip[l](\Mat[W]_{xl} \inp[t] + \Mat[W]_{hl} \pout[t] + \Vet[b]_l)\, ;
\end{equation}
\begin{equation}
 \label{eqLSTM_3}
\Vet[j]^{\left \langle t \right \rangle}=\Sig[j](\Mat[W]_{xj} \inp[t] + \Mat[W]_{hj} \pout[t] + \Vet[b]_j)\, ;
\end{equation}
\begin{equation}
 \label{eqLSTM_4}
\Vet[o]^{\left \langle t \right \rangle}=\Sig[o](\Mat[W]_{xo} \inp[t] + \Mat[W]_{ho} \pout[t] + \Vet[b]_o)\, ;
\end{equation}
\begin{equation}
 \label{eqLSTM_5}
\Mat[C]^{\left \langle t \right \rangle}=(\Mat[C]^{\left \langle t-1 \right \rangle} \odot \Vet[f]^{\left \langle t \right \rangle}) \oplus (\Vet[i]^{\left \langle t \right \rangle} \odot \Vet[j]^{\left \langle t \right \rangle} )\, \text{ and };
\end{equation}
\begin{equation}
 \label{eqLSTM_6}
\out[t]=\Hip(\Mat[C]^{\left \langle t \right \rangle}) \odot \Vet[o]^{\left \langle t \right \rangle} \, \bcom
\end{equation}
where 
$\Mat[W]_{xf}$, $\Mat[W]_{xi}$, $\Mat[W]_{xj}$, and $\Mat[W]_{xo}$ are the weight matrices; 
$\Vet[b]_{f}$, $\Vet[b]_{i}$, $\Vet[b]_{j}$, and $\Vet[b]_{o}$ are the biases; 
$\odot$ and $\oplus$ denote, respectively, the element-wise vectors product and plus; and 
\Sig and \Hip are the activation functions, where \Sig refers to the gates and \Hip is associated with the hidden state and the cell state \cite{jozefowicz2015empirical}. 
Semantically, the architecture means the following:
\begin{itemize}
   \item Forget Gate: decides which information, from the previous hidden state ($\pout[t]$), and the current input ($\inp[t]$), should be kept. The information is transmitted by \Eq\eqref{eqLSTM_1}, ranging from 0 to 1. 
   \painter{Values close to 0 are forgotten, and those close to 1 are kept;}
   
    \item Input Gate: updates the state of the cell state . 
    First, $\pout[t]$ and $\inp[t]$ pass through \Eq\eqref{eqLSTM_3}, that decides which values are to be updated. 
    In parallel, $\pout[t]$ and $\inp[t]$ pass through \Eq\eqref{eqLSTM_2}, which converts values for the range -1 to 1 by \Hip. 
    Then, \Eq\eqref{eqLSTM_2} and \Eq\eqref{eqLSTM_3} are multiplied by $\odot$, so that the \Eq\eqref{eqLSTM_3} is used to decide what information is important to keep from \Eq\eqref{eqLSTM_2}; 

    \item Cell State: connects the outputs of the forget gate to the input gate. 
    Initially, the state of the previous cell $\textbf{C}^{\left \langle t-1 \right \rangle}$ is multiplied by the forget gate. Initially $\textbf{C}^{\left \langle t-1 \right \rangle}=0$. The result is added to the output of the input gate. This procedure leads to the new cell state $\textbf{C}^{\left \langle t \right \rangle}$ (\Eq\eqref{eqLSTM_5});
    
    \item Output Gate: defines the next hidden state $\out[t]$. 
    Initially, the previous hidden state $\pout[t]$ and the current input $\inp[t]$ pass through \Eq\eqref{eqLSTM_4}.
    Then, the state of the newly modified cell ($\textbf{C}^{\left \langle t \right \rangle}$) passes through \Hip. 
    By multiplying the output \Hip by the output from \Eq\eqref{eqLSTM_4} we find out what information the new hidden state should carry. 
   $\textbf{C}^{\left \langle t \right \rangle}$ and $\out[t]$ are transferred to the next stage of time, and the process repeats.    
   \end{itemize}


\subsubsection{\painter{Existing} applications}
\painter{
In poultry farming, variables related animal growth are captured over the time, constituting time series \cite{marquez1995time,palma2007long}. 
Traditional time series forecasting methods often use \emph{autoregressive} models \cite{yang2017robust} as, for example, autoregressive moving average (ARMA) \cite{rojo2004support}, autoregressive integrated moving average (ARIMA) \cite{marquez1995time}, etc. 
However, the literature suggests that \LSTM networks achieve better results for time series, when compared with these and others methods  \cite{abdoli2020comparing,siami2018comparison,jozefowicz2015empirical}.  
}


In agriculture, applications of \LSTM networks are very recent and still limited to a few domains, such as in a greenhouse plan to estimate the temperature, humidity, and $CO_2$ levels \cite{jung2020time}. 
In poultry farming, \LSTM networks have not been applied so far, despite their expected potential. 
They could be used, for example, to predict weight gain of broilers, as the long-term memory feature allows them to trace the accumulative impact of a given action plan, from the first to the last day of production. 
In comparison, a ordinary \RNN would be able to take only the effect of the previous days into account, disregarding the others \cite{Prince,donkoh1989ambient}.

Growth prediction has actually been approached by \cite{johansen2017data}, using \painter{a ARMAX nonlinear} \emph{dynamic neural network} (DNN). However, this model does not allow to assess and trace the impact of different action plans on performance, and variables associated with mortality are not considered. Furthermore, the accumulative long-term memory of \LSTMs leads to a more promising approach to process time-dependent data streams, in comparison to DNNs.

\painter{
Specifically about \FCR[t] estimations, \cite{abreu2020artificial} develop a \ANN model to predict water consumption and body temperature of broilers when subjected to thermal stress. As expected, the \ANN nature prevents to exploit the accumulative impact of distinct action plans over productivity.
Attempting to fulfil this gap in the recent literature, this paper implements and tests a \LSTM network (see \Sect~\ref{sec:main}) that estimates the impact of management variations on the \FCR[t] in \Eq~\eqref{eq1}. Upon validation, this model allows to test different management combinations and quickly sort their best features.
}

\subsection{A search problem appearing from \LSTM results}

When using \LSTMs to estimate \FCR[t], a daily multi-variable action plan ($\inp[t]$) and the previous day output ($\pout[t]$) have to be provided to the network (see \Fig\ref{fig:lstm}). Then, \FCR[t] can be derived from the output $\out[t]$, which is also multi-variable. 
Naturally, every possible combination of inputs leads to a different \FCR[t], and a pertinent question is how to configure an input setup that results into some expected \FCR[t] standard. 

This question can be answered by empirically selecting variable values to compose the \LSTM input. This is actually a trial-and-error strategy very similar to how \painter{specialists} do in practice. 
Under the assumption that the very best \FCR[t] is to be found, it is inevitable to combine every possible input variable value, with all other values, from all other variables. Although this combinatorial task is subject to automation, it is in general associated with exhaustive computing, that grows exponentially with the number of variables to be considered in the problem \cite{woeginger2003exact}. 
As action plans may enlarge their number of variables, exhaustive solutions are limited to reduced setups. 

In this paper, we first implemented an exhaustive search to try to estimate the action plan that would lead to the best \FCR[t]. As expected,  the approach was unable to combine, within useful time, all possible action plan configurations, replicated from the first to the last day of production.

Alternatively, one can replace the exhaustive search by a heuristic that feasibly returns candidate action plans that may, or may not, be the optimal (called the \glob), but a reasonable approximation of it.

\painter{
Although there exist several heuristic methods reported in the literature, such as \emph{Genetic Algorithms} (\GAs), \emph{Simulated annealing},  \emph{Tabu} search, etc. \cite{Gendreau10}, \GAs are of particular interest in this paper. 
\GAs have been one of the most popular heuristic approaches when it comes to solve complex multi-objective and optimisation problems \cite{konak2006multi}. It allows searching simultaneously for different regions of a solution space, which returns a more diversified set of possible solutions for complex problems with non-convex, discontinuous, and multimodal solution spaces \cite{konak2006multi}. 
This coincides with the nature of the application domain exploited in this paper. 
In fact, \GAs have already been applied in poultry farming to search for different options of feed compositions \cite{rikatsih2018adaptive, wijayaningrum2017optimization}, but applications in action plan optimisation has not yet been reported. 
}

\subsection{Genetic Algorithms} \label{subsec:genetic_algo}

A \emph{Genetic Algorithm} is a search method based on the principle of natural evolution: among a population of individuals, the more suitable ones are statistically more likely to reproduce and propagate their \emph{genes} \cite{Goldberg89}. 

Six steps compose a general \GA-based approach:
(i) initial population and restrictions;
(ii) fitness function;
(iii) selection;
(iv) crossover; 
(v) mutation; and
(vi) stopping criteria. 
The \Alg~\ref{alg:alggenetic} \cite{whitley2012genetic} summarizes how they are related. 

\begin{algorithm}[h]
\SetAlgoLined
\Begin{
    defines restrictions\;
    sets population size\;
    generate a initial population\;
\While{Some stopping criterion is not met}{
    compute fitness\;
    apply method for selection\;
    apply method for crossover\;
    apply method for mutation\;
    update the solution\;
    }
return the solution\;
}
 \caption{Pseudo Code - Genetic Algorithm}
 \label{alg:alggenetic}
\end{algorithm}

\subsubsection{Population and restrictions} \label{subsub:populationandrestrictions}
the approach starts by considering a set of individuals (or, biologically speaking, \emph{chromosomes}), that together form a finite \emph{population}, denoted $\pop$, whose size $\Pop[S]$ corresponds to the number of individuals in a \emph{generation}. The larger the population, the more complete is the search for a solution, which increases the chances for it to be a \glob. In contrast, large populations also increase the computational effort to process the search, so that their balance is usually a good measure for successful \GAs. 
In this paper we empirically set $\Pop[S] = 200$ individuals, because it has returned the best cost benefit (accuracy $\times$ execution time) after we tested for 50, 100, 200, 300, and 500 individuals. 
Here, an individual is an action plan and the population $\pop$ is the set of action plans.

Internally, the features of a \emph{chromosome} are represented by a finite chain of \emph{genes}, denoted $\Vet[\Gen]$, of size \Gen[S], which for algorithmic treatment is in this paper modelled by a vector $\Vet[\Gen] = [\gen[1], \cdots, \gen[n]]$.  
Each character $\gen[i],~i=1, \cdots, n$, can be associated with a data domain, such as binary, string, integer, real, etc. 
The way the elements of $\Vet[\Gen]$ are modelled reflects the possibilities for them to be crossed afterwards \cite{mahmudy2013real}. 

For the purposes of this paper, \Vet[\Gen] includes only real and integer numbers.
For example, \Vet[\Gen] = [10, 30.5, 60] exposes a hypothetical action plan with 3 \emph{genes}, i.e., $\Gen[S] = 3$, \painter{where $\gen[1] = 10$ is the day of production, $\gen[2] = 30.5\graus$ describes the temperature and $\gen[3] = 60\%$ describes the humidity}.

Each $\gen[i]$ is limited to well-defined values, or ranges of values, known as \emph{restrictions}. 
There are 3 usual ways to represent restrictions:
(i) equality; 
(ii) inequality; or 
(iii) mixed restriction, that combines the other two \cite{silvaalgoritmos}. 
It is opportune for this paper to focus on inequality restrictions, as they are conducive to intervals.
The model of an inequality restriction is presented in \Eq~\eqref{eq5}: 
\begin{equation}\label{eq5}
A * x \leqslant  b, \, 
\end{equation}
where: $A$ is a matrix of dimension $M$ lines (number of restrictions) by $\Gen[S]$ columns; 
$x$ is the column vector with $\Gen[S]$ elements that contains the variables to be restricted; and 
$b$ is the column vector with $M$ elements.

The following example shows a restriction ($Res$) that bounds the temperature ($T$) to the interval $[30\graus \bcom 35\graus]$, and the humidity ($H$) to $[70\% \bcom 75\%]$. 
\begin{equation}\label{eq8}
Res = 
\begin{bmatrix}
\;\;\;1\;\;\;\;  0 \\ 
-1\;\;\;\;  0 \\ 
\;\;\;0\;\;\;\;  1 \\   \nonumber
\;\;\;0 -1 
\end{bmatrix} 
\begin{bmatrix}
T \\ 
H
\end{bmatrix} \leqslant \begin{bmatrix}
\;\;35 \\
-30 \\
\;\;75 \\
-70 
\end{bmatrix}.
\end{equation}
For this example, $\Gen[S]=2$ and $M = 2 \cdot \Gen[S]$. 

\subsubsection{Fitness and Selection} \label{subsub:fitnes}

\label{p:fit}

a \emph{fitness} function determines the probability for an individual to be selected for reproduction, so that it can propagate its genes to a new descendant. In this paper, fitness refers to \FCR[40] (end of a flock) which is estimated by the three output variables of the \LSTM network.

Based on the fitness value, individuals are \emph{selected}. 
A selection can be implemented by methods such as \emph{Roulette}, \emph{Tournament}, or \emph{Stochastic Uniform} (\SU). 
For the purposes of this paper, the \SU method is shown to be more appropriate, as it chooses several individuals by random sampling and this allows individuals who are weaker in the population (from the point of view of their fitness value) to also have chances to be chosen.

\subsubsection{Crossover} \label{subsubcross} 
after selection, two individuals, or parents, form a child for the next generation, based on their best features, i.e., the child is expected to improve its parents, to some extent \cite{sivanandam2008genetic}. 
\painter{
The crossover method adopted in this paper is the \emph{heuristic crossover} (\HC), which has the characteristic of transferring to the offspring only high fitness regions of the solution space. This leads to quicker convergence and better solutions \cite{pavai2016survey}.
In the poultry literature, the \HC methods has already been tested \cite{wijayaningrum2017optimization,rikatsih2018adaptive} to improve nutrients dosage, and it seems to be sound.
}

Formally, it can be described as in \Eq~\eqref{eq4} \cite{haupt2004practical}. 
\begin{equation}\label{eq4}
\Vet[child] = \Vet[p_2] + \beta* (\Vet[p_1] - \Vet[p_2]) \, .
\end{equation}
That is, the vector $\Vet[child]$ extends its parents (vectors $\Vet[p_1]$ and $\Vet[p_2]$), standing in the path from the worse  ($\Vet[p_1]$) to the best parent ($\Vet[p_2]$). In this case, $\beta $ specifies how far the child is from the parent with best fitness. 
We tested $\beta$ within the interval $[0.1 \bcom 1.2]$ and fixed it in $\beta = 0.6$, which returned better results.

\subsubsection{Mutation} some of the children previously generated by the crossover process may have their genes altered by mutation. This means that a child may receive characteristics not inherited from its parents, which gives more dynamism and diversity to the new population.

Among the existing mutation methods, the \emph{Adaptive Mutation} (\AM) \cite{kumar2010system} has the feature of randomly multiplying some genes by a very low and random value. Upon mutation, the method ensures that genes still respect the bounds imposed by the \emph{restrictions}, \painter{which justifies the use of \AM in this paper}. 

\subsubsection{Stopping Criteria} \label{subsubsec:stopcrite} a \GA has eventually to satisfy a condition that leads it to stop and end the search.
A stopping criteria can be, for example \cite{safe2004stopping}: (i) maximum number of iterations; (ii) maximum execution time; (iii) limit value for the fitness function; (iv) timeout with no changes in the fitness function; (v) restriction violation, etc. In this article, all these stopping criteria were implemented for test. \painter{However, as shown in \Sect~\ref{subsub:expoproposed}, the algorithm stops by reaching the timeout, with no changes in the fitness function.}

For application in poultry houses, we propose next an architecture that integrate \LSTM network as fitness function of the \GA. Together, this combination performs searches for intelligent action plans.

\section{Main Results}\label{sec:main}

\painter{
This section presents our main results. 
Initially, \Sect~\ref{subsec:dataanalysis} formalises the idea of \PCs, and shows how the dataset used in this paper was collected and processed. 
It also presents a mathematical foundation to formally expose action plans, which will support the \LSTM network proposed in \Sect~\ref{subsec:lstm}, and integrated with \GAs in \Sect~\ref{subsec:integration}. 
Finally, physical and logical implementations are presented in \Sect~\ref{sec:implement} in the form of a \emph{Supervisory Control and Data Acquisition} (\SCADA) system and network infrastructure.
}

\subsection{Poultry condominiums} \label{subsec:dataanalysis}

\begin{figure*}[t]
	\psfrag{PH1}[c][c]{\large{\PH[1]}}
    \psfrag{PH2}[c][c]{\large{\PH[2]}} 
    \psfrag{PHN}[c][c]{\large{\PH[n]}}
    \psfrag{ret}[c][c]{\large{\cdots}}
	\centerline{
		\includegraphics[width=\linewidth]{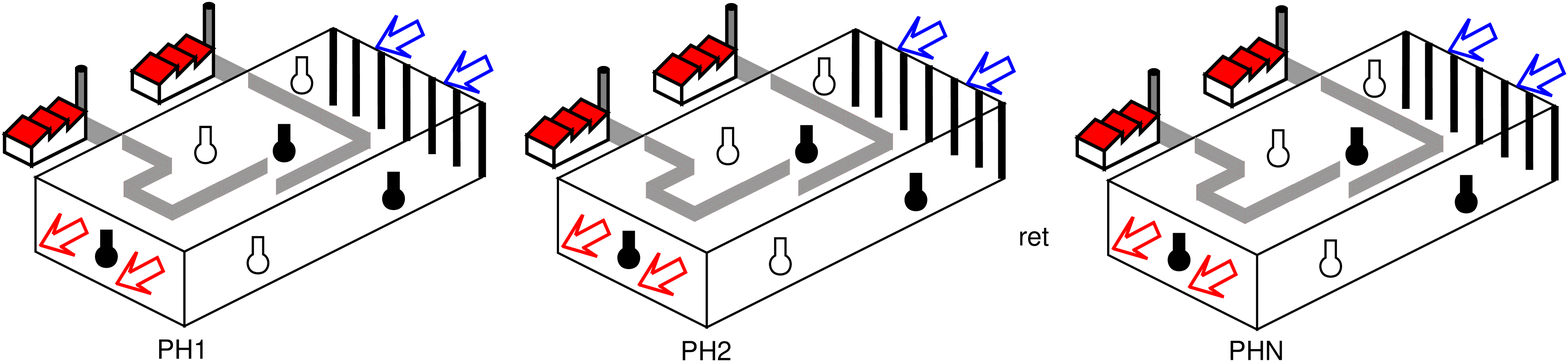}
	}
	\caption{Structure poultry house condominium $\Cond = \{\PH[1] \bcom \cdots \bcom \PH[n]\}$.} 
	\label{fig:Cond}
\end{figure*}


Let a \PC be structured as in \Fig~\ref{fig:Cond} and formalised as the finite set $\Cond = \{\PH[1] \bcom \cdots \bcom \PH[n]\}$ of poultry houses $\PH[i]$, for $1 \leq i \leq n$, that are constructed in proximity and share an isolated, controlled, modular, and escalable infrastructure.
Assume that each \PH[i] is managed by a different specialist so that the action plans they choose may diverge, to some extent. 

Physically, each \PH[i] has an automated feed system, which works continuously from 6 AM until 10 PM, during which broilers can eat according to their needs. It also includes electronic devices that inform the values of daily feed consumption (\dFC[t]), where $\dia$ is the reference day. 
Also automatic scales are positioned along the \house to collect mean daily weight (\MdW[t]) measurements and store them into a database, which is part of the control system infrastructure. 
Each flock takes about $40$ days to be produced, and the final weight (i.e, \MdW[40]) is expected to be of approximately $2.8~\Kg$.

Therefore, for a range $\dia = 1, \cdots, 40$, the temperature (\Min[T], \Avg[T], and \Max[T]) and relative humidity (\Min[H], \Avg[H], and \Max[H]) of the \houses are automatically collect and stored in the database. Complementing for automatic data collection, the specialist also records the \painter{\emph{mortality of day $t$} (\dM[t])}, which can be alternatively exposed as the \emph{number of live birds} (\NlB[t]), such that: 
  \begin{equation}\label{eq:b}
        \NlB[t] =  |\Flock| - \sum_{t =1}^{p}~\dM[t] \, . 
    \end{equation}
That is, \NlB[t] is the difference between the \emph{initial amount of birds} in a flock ($|\Flock|$), and the accumulated mortality recorded by the \painter{specialist} during a period $p$.

\subsubsection{Data acquisition}\label{sec:aquisition}
for the purpose of this paper, the dataset was constructed by observing a poultry condominium composed of three \houses, which were used for collecting daily data sampling. The birds in the three \houses were only male, \emph{Ross} broilers \cite{bross:Online}. 
The data acquisition was made between August 2019 and March 2020, alternating therefore different seasonal features. 
A total of 12 flocks were observed (\painter{i.e., 12 samples}), the same as used in related literature \cite{johansen2017data}, which has shown to be quite sufficient for the intended analysis.

Table~\ref{tab:variable} summarises and denotes the main variables collected and used in the experiments that follow. 
Input and output variables have been formalised so far, while \emph{output normalised variables} will be introduced in the following section \Sect~\ref{subsubsec:data_normalization}. 
\painter{
\emph{Input variables} correspond to the action plan for day $\dia$, for $\dia = 1,\cdots,40$), while \emph{output variables} expose the variables influenced by this action plan. 
The selection algorithm \emph{Greedy Stepwise} \cite{gevrey2003review} has been applied to confirm the relevance of the input variables over the output. According to the tests performed, all input variables influence the output variables.}

\begin{table}[h]
\centering
\begin{tabular}{r|r|r} \hline
{\bf Input variables}   & \painter{Unit} & {\bf Notation} \\ \hline
day {}{}                & $\dia = 1,\cdots,40$ &    \dia \\ \hline
minimum temperature     & {[}\graus{]} &    \Min[T] \\
mean temperature        & {[}\graus{]} &    \Avg[T] \\
maximum temperature     & {[}\graus{]} &   \Max[T] \\ \hline
minimum humidity            & {[}\%{]} &   \Min[H] \\
mean humidity               & {[}\%{]} &   \Avg[H] \\
maximum humidity            & {[}\%{]} &   \Max[H] \\ \hline
{\bf Output variables}              &  &   {\bf Notation} \\ \hline
mean daily weigh        & {[}\grama{]} &      \MdW[t] \\ 
daily feed consumption  & {[}\Kg{]} &      \dFC[t] \\ 
number of living birds  & {[}birds{]} &   \NlB[t] \\ 
daily mortality         & {[}birds{]} &   \dM[t] \\ \hline
{\bf Output variables normalized}                               &  & {\bf Notation} \\ \hline
daily feed consumption per bird         & {[}$\Kg/bird${]} &   \dFCpB[t] \\ 
number of living birds per area           & {[}$bird/\meter^{2}${]} & \NlBpA[t] \\ 
daily mortality per area                & {[}$bird/\meter^{2}${]} & \dMpA[t] \\ \hline
\end{tabular}  
\caption{Variables to be considered in our experiments. 
}
\label{tab:variable}
\end{table}

A daily \painter{measurement} was collected for each of the $7$ input variables in Table~\ref{tab:variable}, totalling $280$ \painter{measurements} ($7 \times 40$ days of the flock) \painter{in one sample}. For the $3$ output variables, the daily \painter{measurement} for \MdW[t], \dFC[t], and \NlB[t] correspond to $120$ \painter{measurements} ($3 \times 40$), \painter{also for one sample}. 
Therefore, \painter{as there were $3$ \houses, each one provided $4$ samples to the total of $12$ samples, which corresponds to a input data matrix of $12 \times 280$, while the corresponding output data form a matrix sized $12 \times 120$.}


\subsubsection{Data normalisation} \label{subsubsec:data_normalization}
the observed \houses have two different areas, $\A_{\PH[1]}$ and $\A_{\PH[2]}$. 
$\A_{\PH[1]}$ is $150~\meter$ long and $16~\meter$ wide, totalling $2400~\mtwo$, and it can host about $34800$ birds. 
$\A_{\PH[2]}$ is $150~\meter$ long by $12~\meter$ wide, with a floor area of $1800~\mtwo$ that accommodates $26500$ birds. 
Thus, \dM[t] and \NlB[t] can be normalised by the area $\A_{\PH[i]}$, such that: 
    \begin{equation}\label{eq:MpA}
        \dMpA[t] =  \frac{\dM[t]}{\A_{\PH[i]}}\, ; \text{ and } 
    \end{equation}
    \begin{equation}\label{eq:bpa}
        \NlBpA[t] =  \frac{\NlB[t]}{\A_{\PH[i]}} \, \cdot
    \end{equation}
Also, \dFC[t] can be normalised by \NlB[t], resulting in \emph{feed consumption per bird} (\dFCpB[t]), such that:
    \begin{equation}\label{eq:cpb1}
        \dFCpB[t] = \frac{\dFC[t]}{\NlB[t]}\, \cdot
    \end{equation}
    
Remark that, in this way, \FCR[t] can be calculated by modifying the \Eq~\eqref{eq1}, additionally pondering:
(i) the number of living birds per area $\A_{\PH[i]}$ (\NlBpA[t]); and 
(ii) the daily feed consumption by the number of living birds (\dFCpB[t]).
\Eq~\eqref{eq3} formalises this idea:
\begin{equation}
 \label{eq3}
\FCR[t] = 1000 ~\frac{\dFCpB[t]}{\NlBpA[t]~\MdW[t]}~\NlBpA[t] \, ,
\end{equation}
where the constant $1000$ compensates the effect of the weight conversion from \Kg to \grama in \dFCpB[t].

The collected data can be statistically assessed as follows.

\subsubsection{Exploratory data analysis} 
\emph{mean} (\mean) and \emph{standard deviation} (\dev) are two essential metrics to be considered in this paper. 
For the set of variables in Table~\ref{tab:variable}, collected during the 12 observed flocks, \mean and \dev are summarized as in \Fig~\ref{fig:datacolec}. 

\def\scaleIMG{0.158}
\begin{figure*}[t]
\psfrag{Axisx}[c][c]{\scriptsize{day [t]}}
    \centering
	    \subfigure[][{Input variables: minimum temperature  }]{
		\psfrag{Axisy}[c][c]{\scriptsize{\Min[T] [$\graus$]}}
		\includegraphics[scale = \scaleIMG]{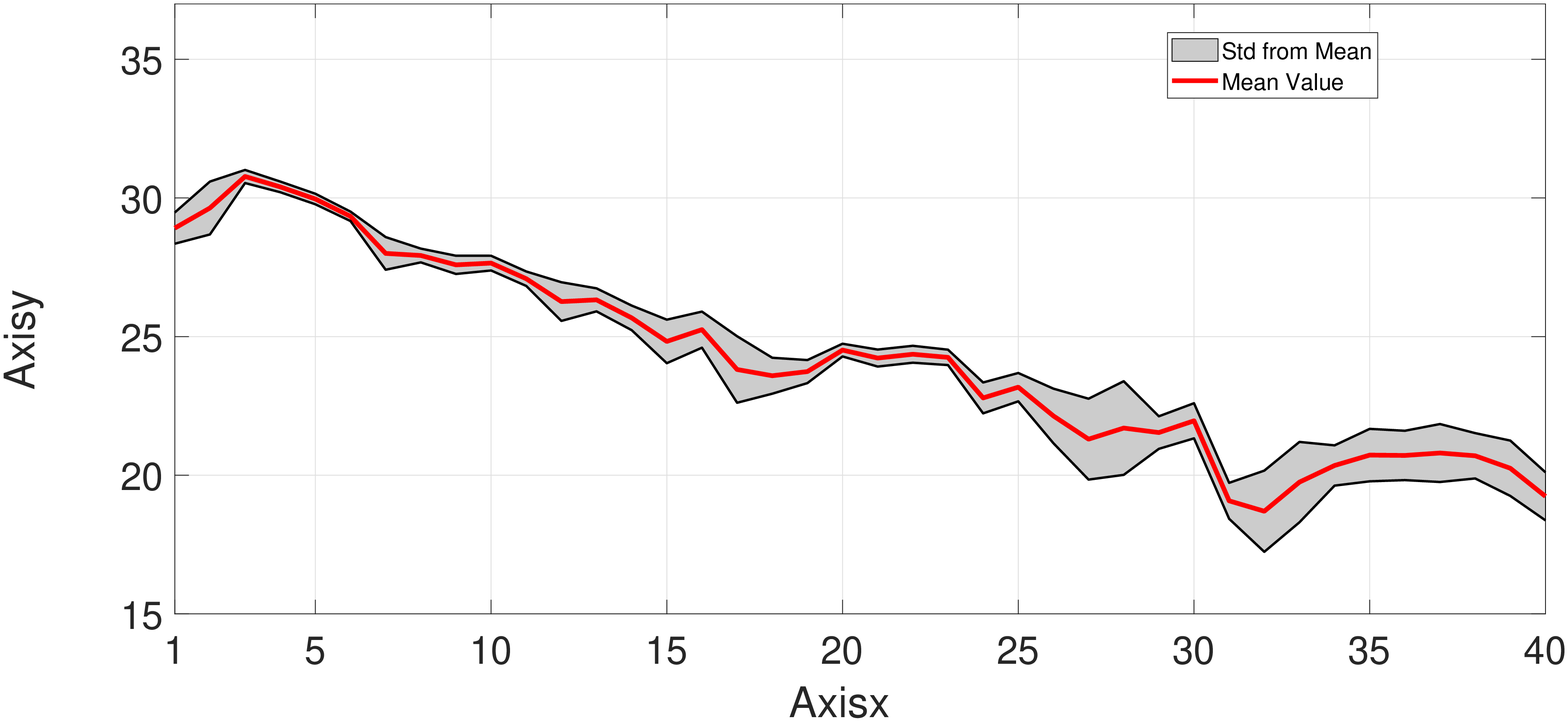}
		\label{fig:temp_min}
	}
		\subfigure[][{Input variables: mean temperature}]{
		\psfrag{Axisy}[c][c]{\scriptsize{ \Avg[T] [$\graus$]}}
		\includegraphics[scale = \scaleIMG]{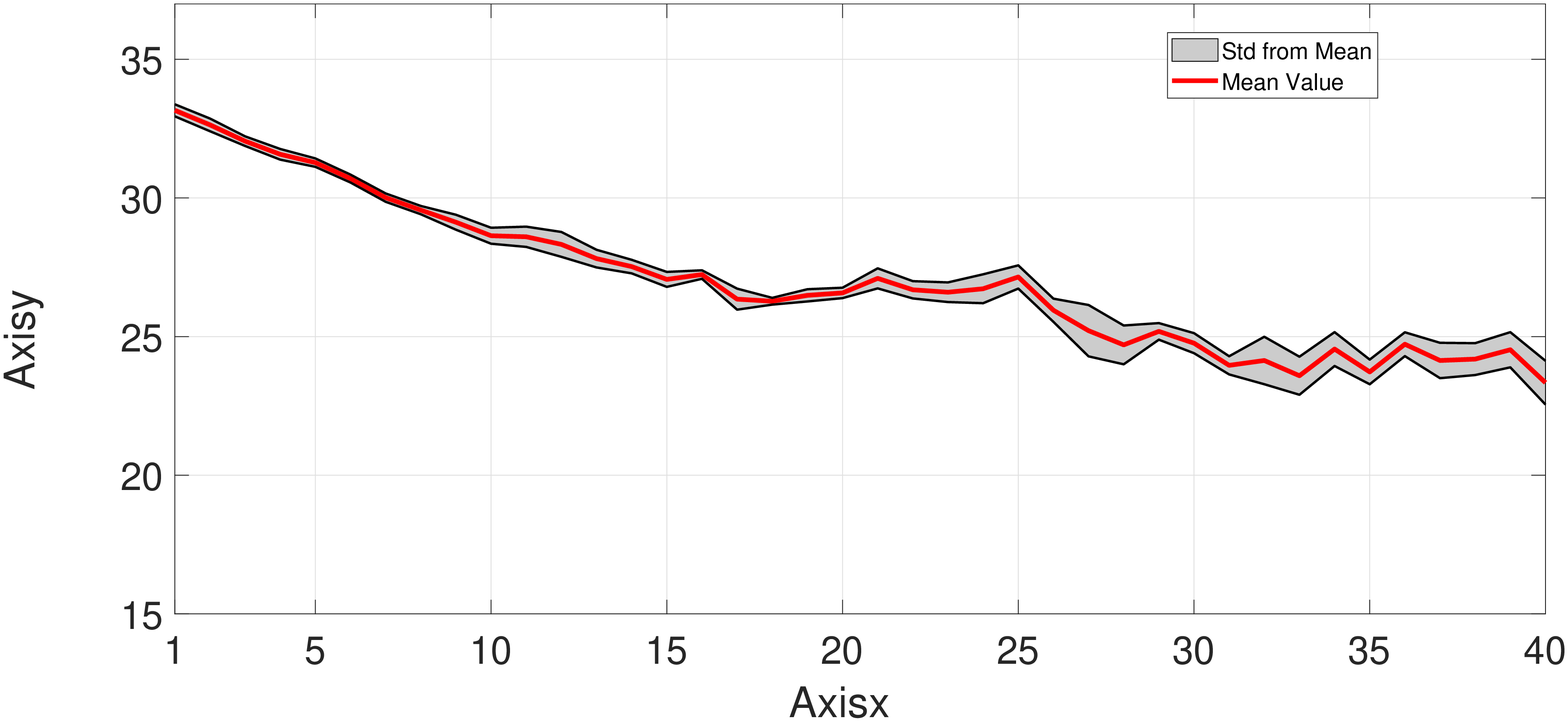}
		\label{fig:temp_med}
	}
		\subfigure[][{Input variables: maximum temperature }]{
		\psfrag{Axisy}[c][c]{\scriptsize{\Max[T] [$\graus$]}}
		\includegraphics[scale = \scaleIMG]{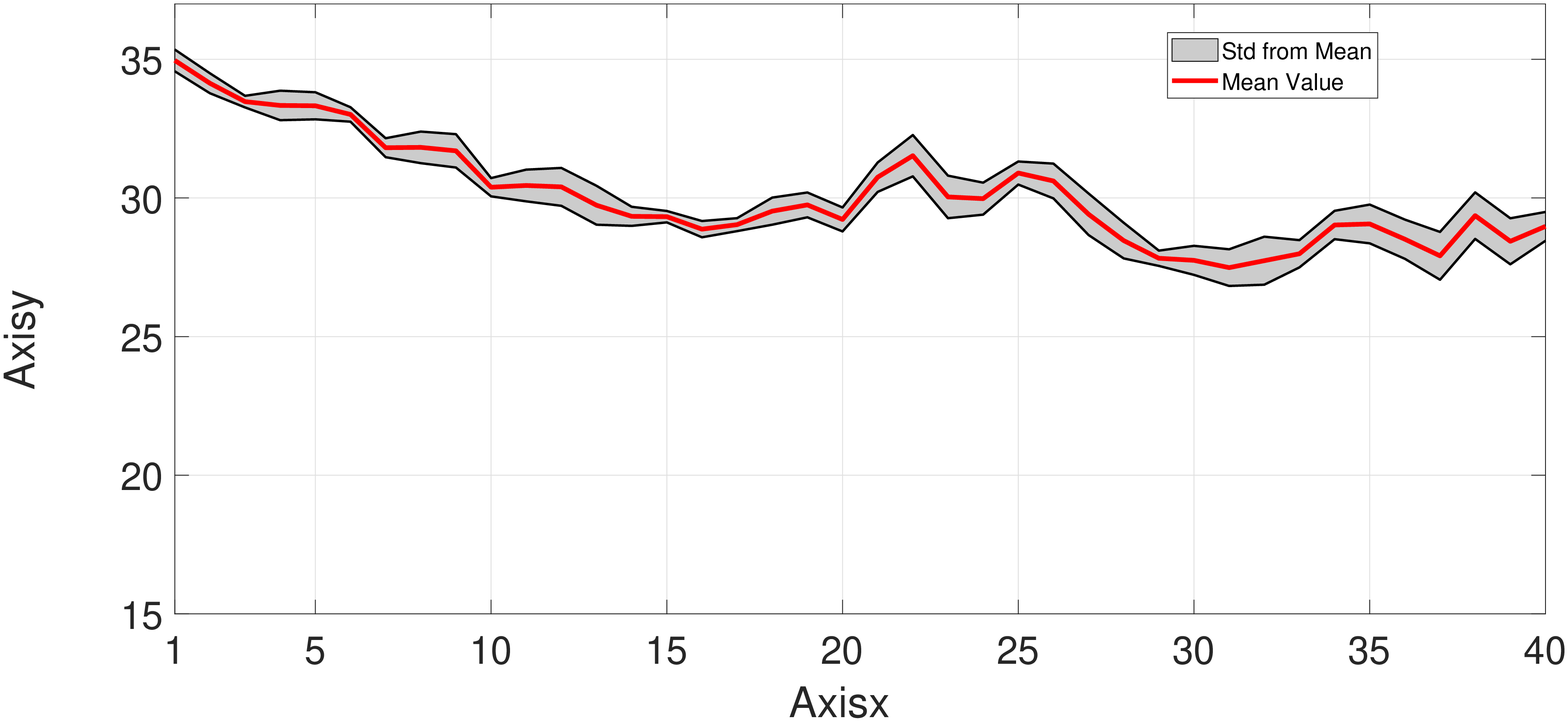}
		\label{fig:temp_max}
	}
	\subfigure[][{Input variables: minimum humidity }]{
		\psfrag{Axisy}[c][c]{\scriptsize{\Min[H] [\%]}}
		\includegraphics[scale = \scaleIMG]{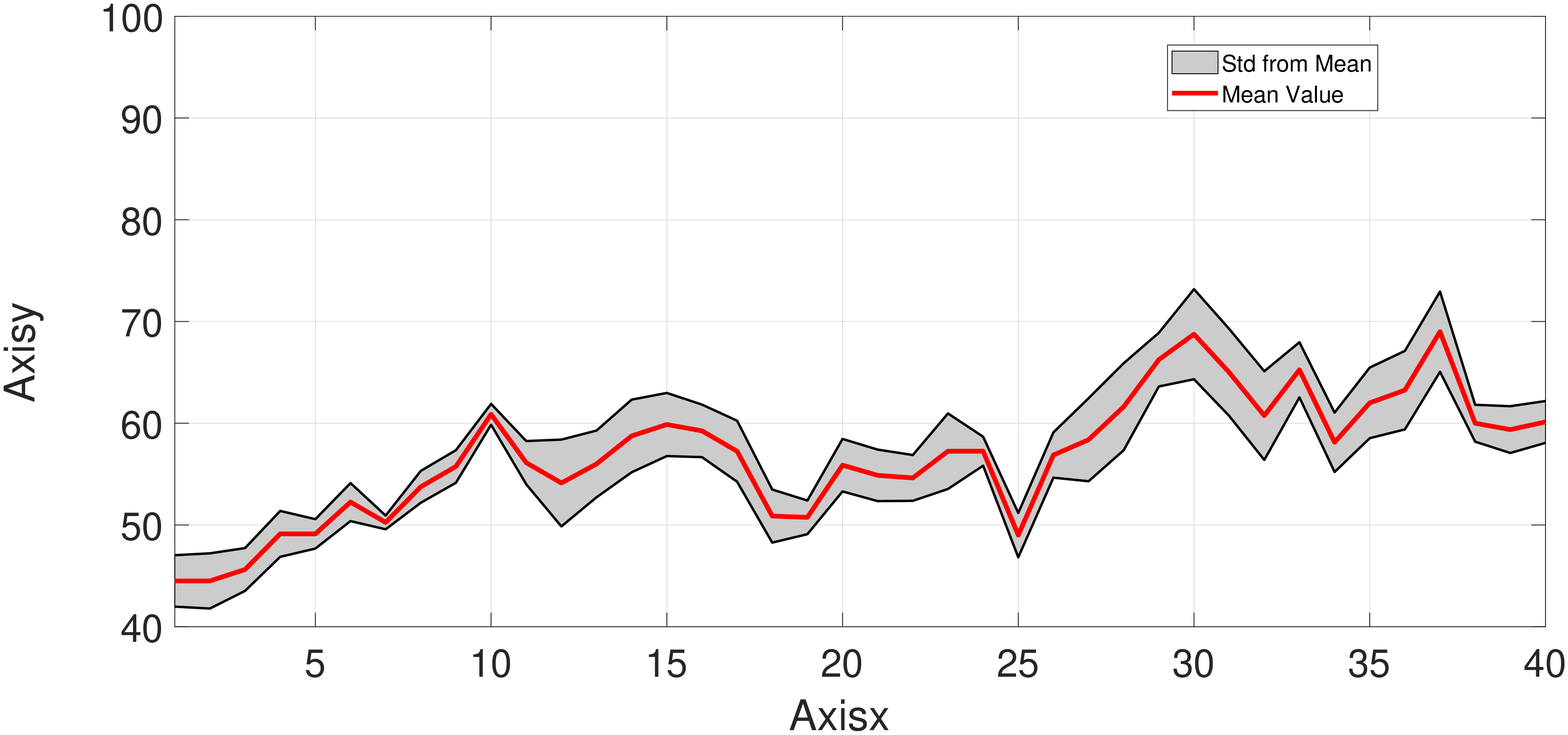}
		\label{fig:hum_min}
	}
	\subfigure[][{Input variables: mean humidity }]{
		\psfrag{Axisy}[c][c]{\scriptsize{\Avg[H] [\%]}}
		\includegraphics[scale = \scaleIMG]{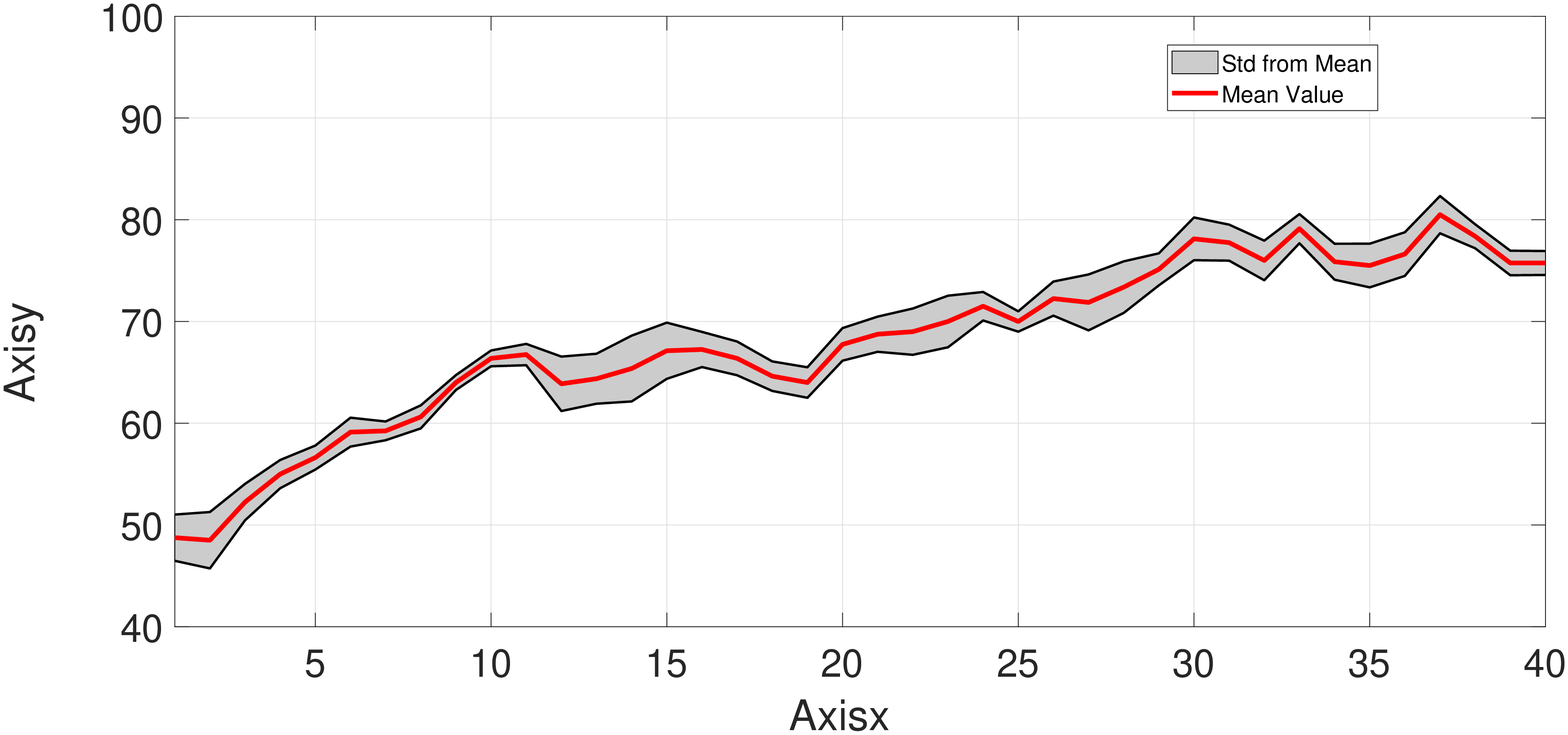}
		\label{fig:hum_med}
	}
	\subfigure[][{Input variables: maximum humidity }]{
		\psfrag{Axisy}[c][c]{\scriptsize{\Max[H] [\%]}}
		\includegraphics[scale = \scaleIMG]{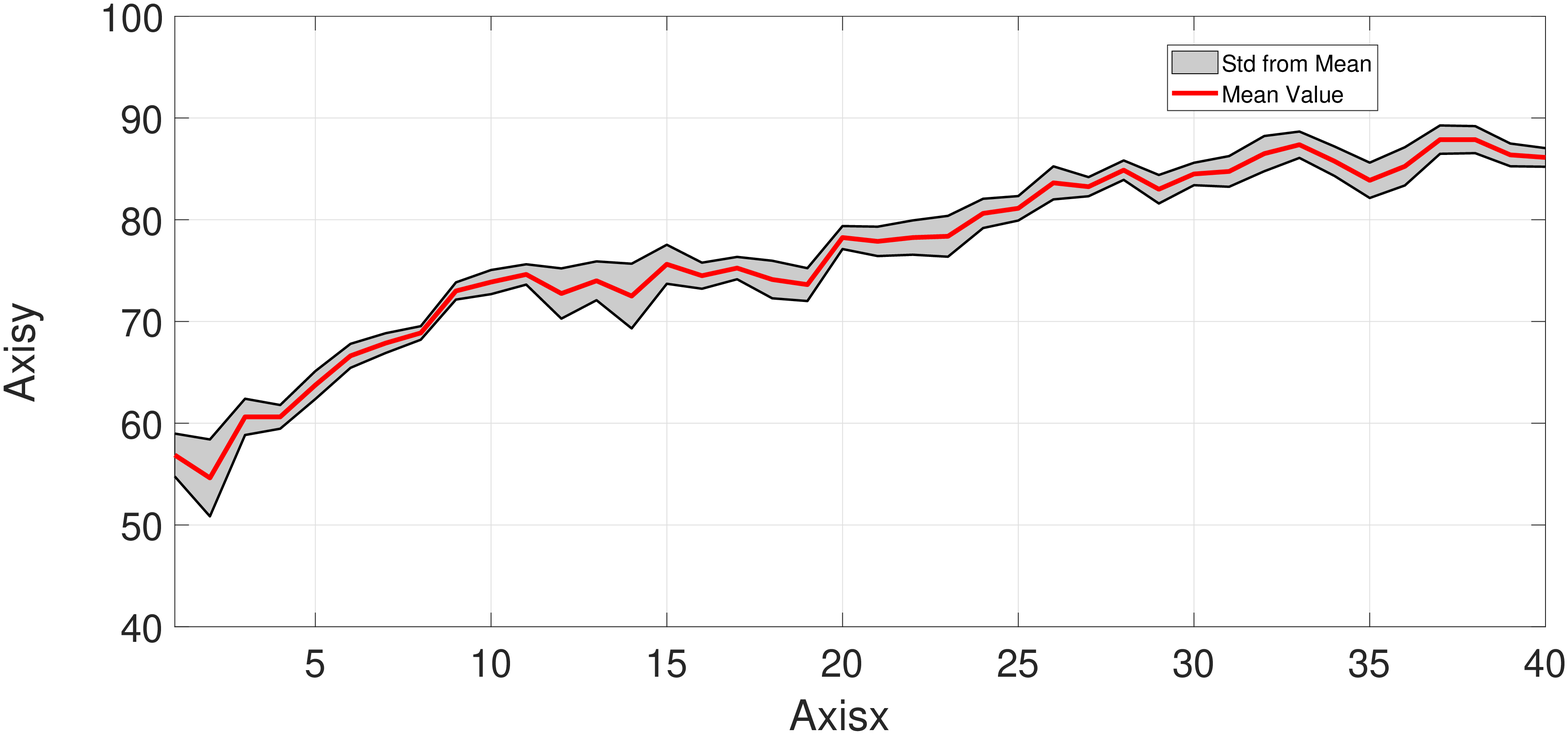}
		\label{fig:hum_max}
	}
	\subfigure[][{Output variables: mean daily weigh }]{
		\psfrag{Axisy}[c][c]{\scriptsize{$\MdW[t]$ $[\grama]$}}
		\includegraphics[scale = \scaleIMG]{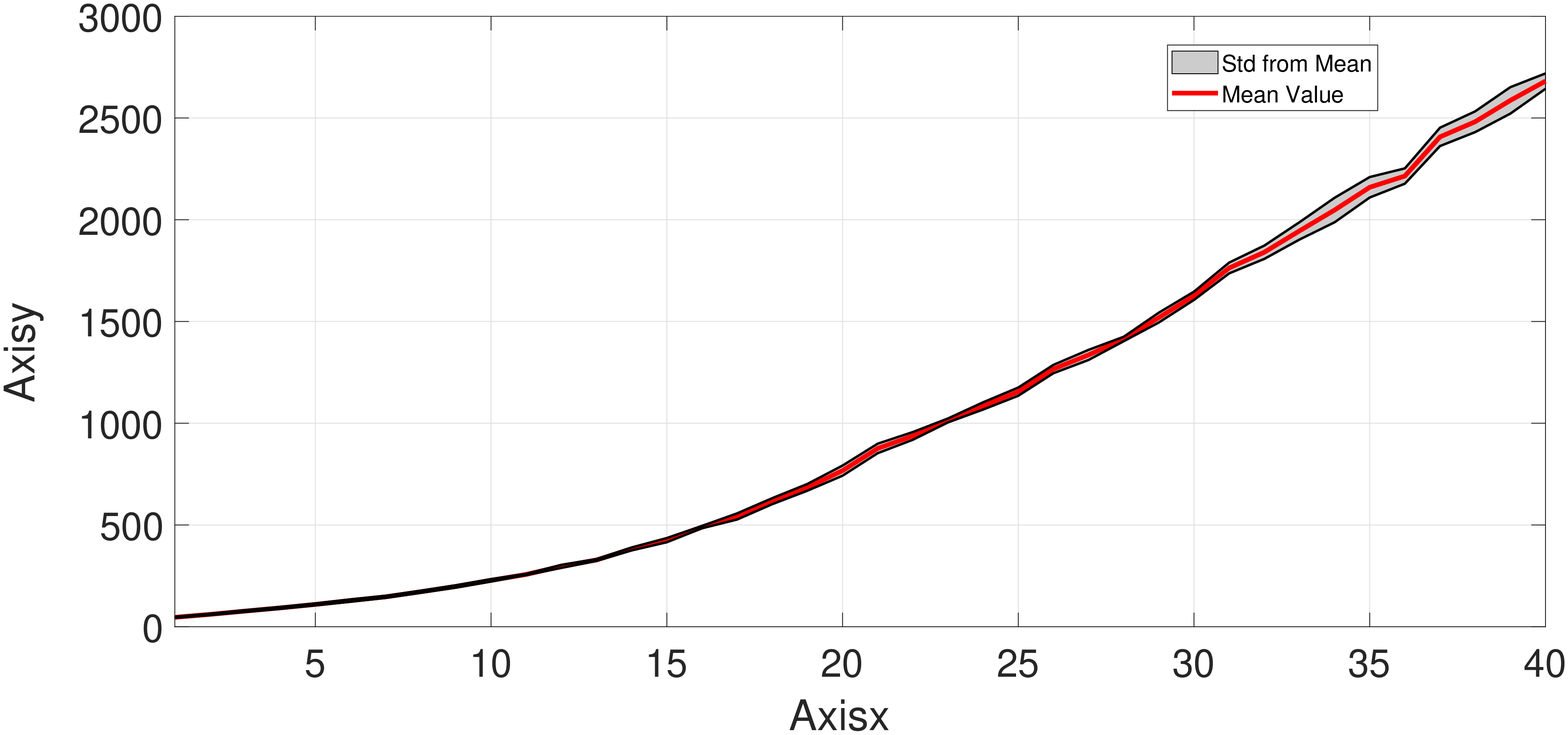}
		\label{fig:peso}
	}
	\subfigure[][{Output variables normalized: daily feed consumption per bird}]{
		\psfrag{Axisy}[c][c]{\scriptsize{$\dFCpB[t]$ $[\Kg/bird]$}}
		\includegraphics[scale = \scaleIMG]{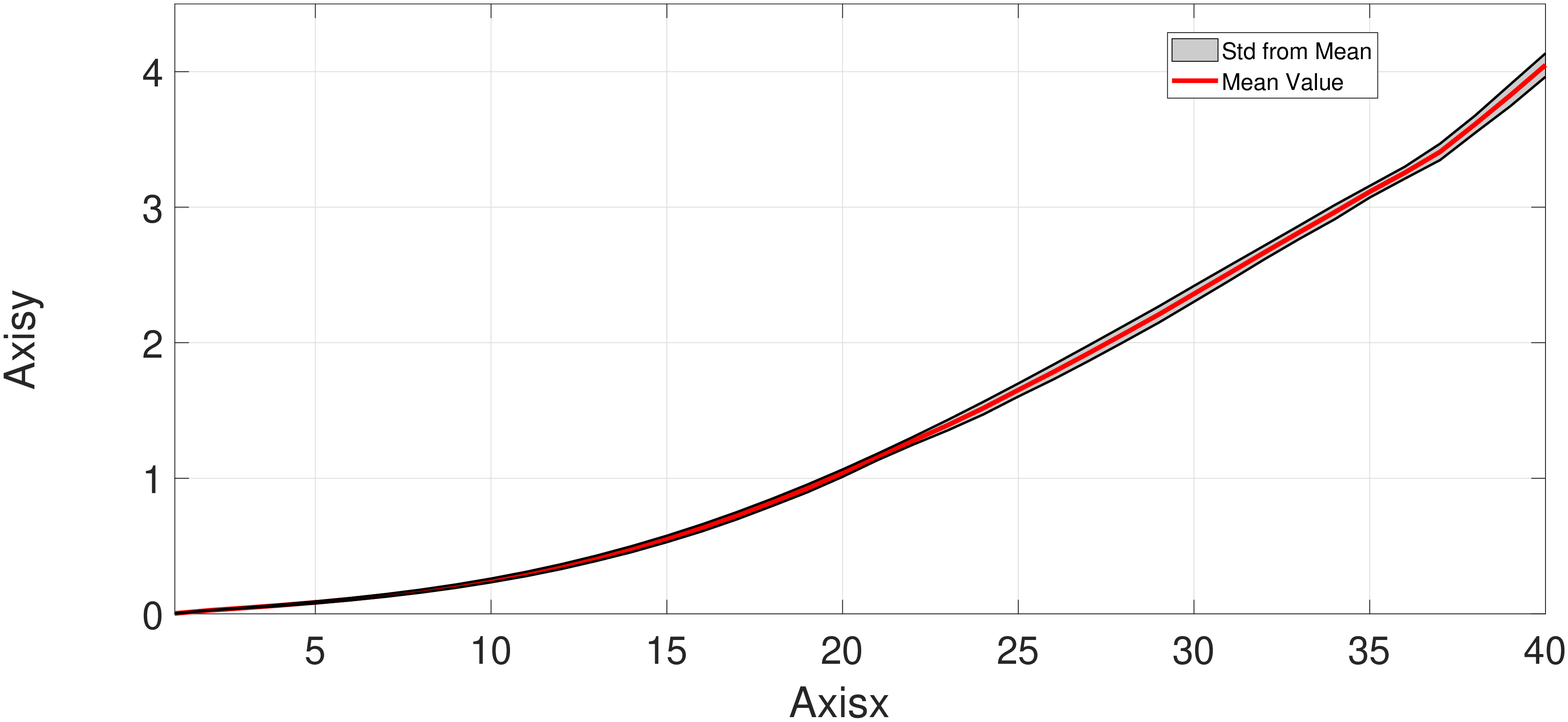}
		\label{fig:consumo}
	}
	\subfigure[][{Output variables normalized: daily mortality per area}]{
		\psfrag{Axisy}[c][c]{\scriptsize{$\dMpA[t]$ $[bird/\meter^{2}]$}}
		\includegraphics[scale = \scaleIMG]{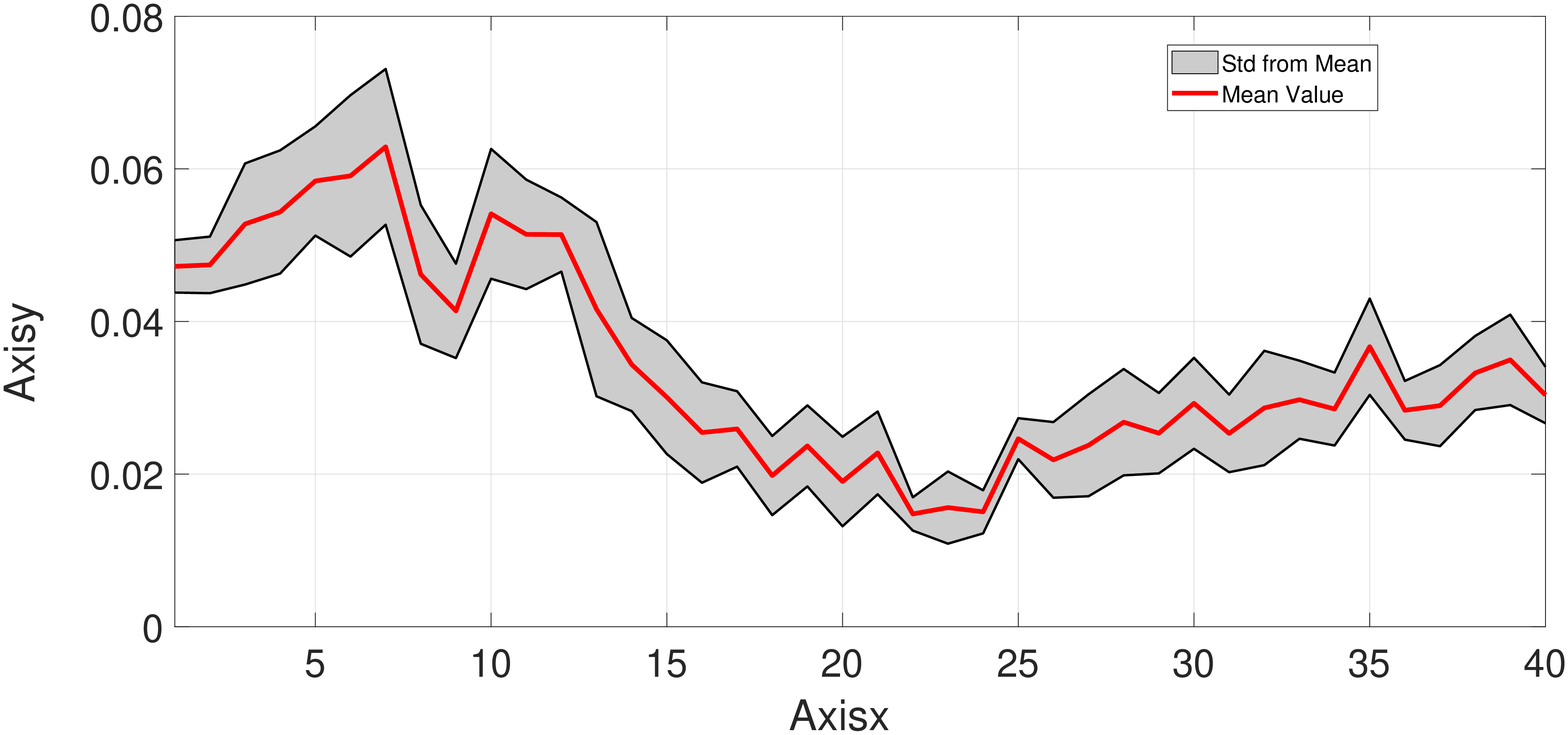}
		\label{fig:mortes}
	}
    \caption{Statistical treatment of the collected variables.}
    \label{fig:datacolec}
\end{figure*}

One can observe a strong positive linear correlation (coefficient $0.9996$, $\cong 1$) between \Mean[{\dFCpB[t]}] in \Fig~\ref{fig:consumo} and \Mean[{\MdW[t]}] in \Fig~\ref{fig:peso}, similarly to what is observed in \cite{johansen2017data}. 
\painter{See in \Fig~\ref{fig:peso} that \Dev[{\MdW[t]}] remains without significant variation until approximately day $20$.}




\painter{
Furthermore, \Mean[{\MdW[t]}] has a negative linear correlation (coefficient $- 0.90$) with \Mean[{\Avg[T]}] (\Fig~\ref{fig:temp_med}), which, in its turn, also show no significant variations of \Dev[{\Avg[T]}] (\Fig~\ref{fig:temp_med}) around the average by the same period of twenty days. 
We believe that, due to climate homogeneity, the same \Avg[T] standard was set to all \houses during this period. 
Another possibility is that the homogeneity results from the reduced space into which the birds are subject during their first twenty days of life. After this period, they gradually occupy the entire space of the \house, which equivalently complexifies the temperature control, possibly increasing \Dev[{\Avg[T]}]. 
}




\Fig~\ref{fig:mortes} suggests that \Mean[{\dMpA[t]}] is seasonal and more concentrated during the initial and final lifespan. 
From day 1 to day 10, a higher mortality rate is noticed, which we believe is caused by periodic interventions of the specialist who manually eliminates birds that, in his opinion, are not healthy enough to return a good \FCR[t] and could undermine the entire \FCR[t] of the flock. 
\painter{
From the day 25 forward, mortality starts a new upward trend, intensifying it over the remaining days. 
A closer inspection on \Mean[{\Min[T]}] (\Fig~\ref{fig:temp_min}) and \Mean[{\Max[T]}] (\Fig~\ref{fig:temp_max}) reveals that this may be caused by thermal stress provoked by temperature spikes.} 



Also remark that \Mean[{\Avg[H]}] (\Fig~\ref{fig:hum_med}) shows a negative linear correlation (coefficient $- 0.96$) with \Mean[{\Avg[T]}] in \Fig~\ref{fig:temp_med}. 
In fact, \Avg[H] (\Fig~\ref{fig:hum_med}), \Min[H] (\Fig~\ref{fig:hum_min}), and \Max[H] (\Fig~\ref{fig:hum_max}), concentrate around the average and they have an increasing rate until the day 10. 
This should be also due to the small confinement space, which favours the climate homogeneity.

From the day 10 forward, there is an increase in the standard deviation (\Dev[{\Min[H]}], \Dev[{\Avg[H]}], and  \Dev[{\Max[H]}]).
One possibility is that this is caused by manual interventions by the specialist who opens and closes the small side windows used for air intake and renewal. 
Another possible cause is sudden climate variations outside the \house, for example with long rainy season, or cold weather intake. 
Particularly in cold mornings and winter-like seasons, internal and external climate conditions tend to contrast significantly.



\subsubsection{Reliability analysis}
now, we investigate how representative the collected dataset is \wrt the generic conditions \houses. 
By using \emph{statistical inference} on the average of temperature and humidity, we derive a weekly confidence interval, using the \emph{Student’s t-probabilistic} distribution \cite{hooft2009probability}. 

\begin{figure}[!htb]
\psfrag{Value}[c][c]{Value}
\psfrag{Frequency}[c][c]{\scriptsize{Frequency}}
    \centering
	\includegraphics[width=7.4cm]{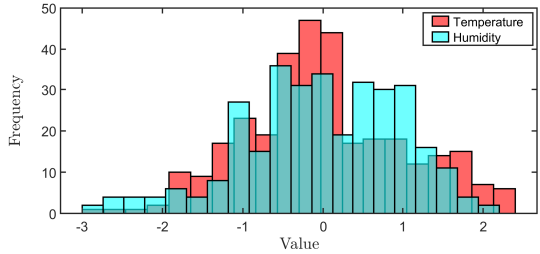}
	 \centering
	\caption{Statistical relevance of the collected dataset.}
\label{fig:histograms}
\end{figure}

The histograms in \Fig~\ref{fig:histograms} show that the mean temperature (\Avg[T]) and the mean humidity (\Avg[H]) have a approximately normal distribution for the period of one week.
For representation, the \Avg[T] and \Avg[H] data were normalized by \emph{Z-Score} \cite{gravetter2020essentials}.

Considering a 95 \% confidence level, the results of the probabilistic outcomes are shown in the Table~\ref{tab:confi_interval}. In practice, \Mean[{\Avg[T]}] and \Mean[{\Avg[H]}] centralise the weekly interval, plus a small deviation upwards and downwards.

\begin{table}[h]
\centering
\begin{tabular}{c|cc}
\hline
\textbf{Week} & \Avg[T] [$^{\circ}$C] & \Avg[H] [\%] \\ \hline 
1 & {[30.36, 32.47}{]} & {[49.46, 60.18}{]} \\ \hline
2 & {[27.19, 29.07}{]} & {[60.42, 70.34}{]} \\ \hline
3 & {[25.93, 27.30}{]} & {[62.94, 71.76}{]} \\ \hline
4 & {[23.95, 26.92}{]} & {[68.86, 78.77}{]} \\ \hline
5 & {[22.77, 25.65}{]} & {[72.98, 81.43}{]} \\ \hline
6 & {[22.72, 25.63}{]} & {[73.57, 81.22}{]} \\ \hline
\end{tabular}
\caption{Confidence Interval.}
\label{tab:confi_interval}
\end{table}

From this analysis, we conclude that weekly true values for \Avg[T] and \Avg[H] are 95\% probable to be within the intervals in Table~\ref{tab:confi_interval}. 
Also, the intervals have a low amplitude, which strengthens the hypothesis that our dataset is acceptably representative.

\subsection{Proposed \LSTM Neural Network}\label{subsec:lstm}

\begin{figure}[!htb]
\psfrag{dia}[r][l]{\scriptsize{\dia}}
\psfrag{temperatura_minima}[r][l]{\scriptsize{\Min[T]}}
\psfrag{temperatura_media}[r][l]{\scriptsize{\Avg[T]}}
\psfrag{temperatura_maxima}[r][l]{\scriptsize{\Max[T]}}
\psfrag{humidade_minima}[r][l]{\scriptsize{\Min[H]}}
\psfrag{humidade_media}[r][l]{\scriptsize{\Avg[H]}}
\psfrag{humidade_maxima}[r][l]{\scriptsize{\Max[H]}}
\psfrag{peso_anterior}[r][l]{\scriptsize{\pMdW}}
\psfrag{consu_anterior}[r][l]{\scriptsize{\pdFCpB}}
\psfrag{vivos_anterior}[r][l]{\scriptsize{\pNlBpA}}
\psfrag{Action}[][][1][90]{\scriptsize{\textbf{Action Plan}}}
\psfrag{Last}[][][1][90]{\scriptsize{\textbf{Last Day}}}
\psfrag{peso}[c][c]{\scriptsize{\MdW[t]}}
\psfrag{unit_1}[c][c]{\scriptsize{[g]}}
\psfrag{consu}[c][c]{\scriptsize{\dFCpB[t]}}
\psfrag{unit_2}[c][c]{\scriptsize{[\Kg / bird]}}
\psfrag{vivos}[c][c]{\scriptsize{\NlBpA[t]}}
\psfrag{unit_3}[c][c]{\scriptsize{[bird / \mtwo]}}
\psfrag{CA}[c][c]{\scriptsize{\FCR[t]}}
\psfrag{unit_4}[c][c]{\scriptsize{[\Kg/}}
\psfrag{unit_5}[c][c]{\scriptsize{(g bird)]}}
\psfrag{EQ3}[][][1][90]{\footnotesize\Eq~\eqref{eq3}}
\psfrag{LSTM}[][][1][90]{\footnotesize{LSTM Neural Network}}
    \centering
	\includegraphics[width=7cm]{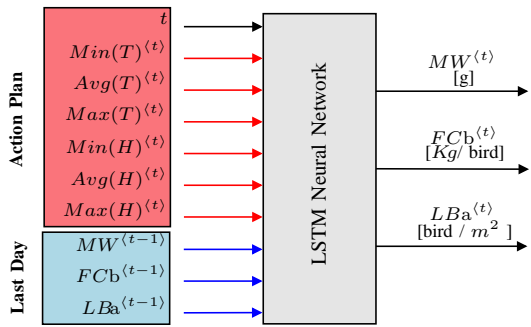} 
	\caption{Proposed \LSTM network architecture.}
\label{fig:arquitetura_rede}
\end{figure}

Now, we exploit our dataset in order to support decision making in poultry farming daily management.  
The first proposal is based on the \LSTM neural network shown in \Fig~\ref{fig:arquitetura_rede}. 
It takes as input the day under analysis (\dia); 
temperature (\Min[T], \Avg[T], and \Max[T]);
humidity (\Min[H], \Avg[H], and \Max[H]);
the mean daily weigh of the previous day (\pMdW); 
the daily feed consumption per bird in the previous day (\pdFCpB); and 
the number of living birds per area in the previous day (\pNlBpA). 

From now forward, the \LSTM inputs and outputs will be referred as vectors, as in \Fig~\ref{fig:lstm}. That is:
\begin{equation}
 \label{eq:atribuition1}
\inp[t]=[\dia,\Min[T], \cdots, \Max[H]] \, ;
\end{equation}
\begin{equation}
 \label{eq:atribuition2}
\pout[t]=[\pMdW, \pdFCpB, \pNlBpA] \, ; 
\end{equation}
\begin{equation}
 \label{eq:atribuition3}
\out[t]=[\MdW[t], \dFCpB[t], \NlBpA[t]] \, .
\end{equation}

Vector 
$\inp[t]$ carries the input action plan for day $\dia$;
$\pout[t]$ is the input vector with the previous days outputs; and
$\out[t]$ is the output vector for day $\dia$. 
By concatenating $\inp[t]$ and $\pout[t]$ (denoted $\inp[t]{^{\frown}} \pout[t]$), and by applying it as the neural network input, we obtain a prediction for $\out[t]$.  

\subsubsection{An efficient modular proposal}
In preliminary tests, we approached the dataset monolithically, i.e., considering a single model for the 40 days. 
However, a modular strategy has shown to be more accurate \painter{and consistent with practice. 
In fact, weekly updates in the setup of \houses are quite usual. Periodically, the specialist has to adjust the height of drinking fountains, the intensity of ventilation and lighting, increase the locomotion area, etc. \cite{bross:Online}. Therefore, a modular design of \LSTM is more likely to capture every new physical setup. 
}

Consider partitioning the model in \fig~\ref{fig:arquitetura_rede} into 6 replicas (denoted by $\wm_{\week=1} \bcom \cdots \bcom \wm_{\week=6}$), so that each partition models a specific week of production.
Five partitions ($\wm_{\week=1},\cdots,\wm_{\week=5}$) comprise seven days (denoted by $\week_S=7$, with $\week_S$ representing the size of the week) of life each, and the other partition ($\wm_{\week=6}$) maps the remaining five days ($\week_S=5$), completing the same 40 days of life. 

Similarly, the entire dataset (12x280 input and 12x120 output data) is also partitioned into 6 smaller datasets: 
5 datasets of dimension 12x49 for input and 12x21 for output data; and 
1 dataset of 12x35 for input and 12x15 for output data. 
Besides, from the 12 collected samples, 10 were allocated for network training, and 2 for test, the same setup as in \cite{johansen2017data}.

\subsubsection{Data pre-processing} \label{data_pre_processing}
before execution, we normalise the \LSTM inputs to the interval $[0,1]$ in order to improve stability and performance of the deep learning model. 
\Eq~\eqref{eq6} formalises a function \Norm that returns the normalised version of a vector $\Vet[\kap]$, based on its upper and lower bounds, such that:
\begin{equation} \label{eq6}
\Norm(\Vet[\kap], \Vet[\maxi], \Vet[\mini]) =  \frac{\Vet[\kap] - \Vet[\mini]}{\Vet[\maxi] - \Vet[\mini]} \, \bcom
\end{equation}
where $\Vet[\maxi]$ and $\Vet[\mini]$ are upper and lower bounds of $\Vet[\kap]$, i.e., $\Vet[\mini] \leqslant  \Vet[\kap] \leqslant  \Vet[\maxi]$. 
Let, for example, 
$\Vet[\kap] = [24,26,27]$ be 3 temperature measurements collected from the a \house floor. 
Consider that 
$\Vet[\mini]=[23,22,25]$, and 
$\Vet[\maxi]=[28,27,30]$ are the corresponding lower and upper bounds for $\Vet[\kap]$. 
Then, the temperature ($T$) in the position $1$ ($T_1$), for instance, varies from 23 to 28, such that $\Norm([24], [28], [23])= 0.2$, and 
$\Norm(\Vet[\kap],\: \Vet[\maxi],\: \Vet[\mini]) = [0.2,0.8,0.4]$.

The inverse mapping, denoted \DNorm (\Eq~\eqref{eq7}), transforms the \LSTM outputs back to their original domain.
\begin{align}\label{eq7}
\centering
 & \DNorm(\Norm(\Vet[\kap], \Vet[\maxi], \Vet[\mini]))& \nonumber\\ 
 & =& \nonumber \\
 & \Norm(\Vet[\kap], \Vet[\maxi], \Vet[\mini]) (\Vet[\maxi] - \Vet[\mini]) + \Vet[\mini]. \, &  
\end{align}
This normalisation is also called min-max scale \cite{normalizations}.

\subsubsection{\painter{Hyper-parameters settings for training the \LSTM model}}
the task of tuning optimal hyper-parameters for the \LSTM model can be approached in different ways. 
In this paper, we use an empirical-guided random search over the grid space.

{\painterX 
Our focus was to adjust seven important hyperparameters: 
(i) number of epochs;
(ii) learning rate;
(iii) number of layers;
(iv) number of neurons;
(v) initialisation of weights;
(vi) activation functions;
(vii) regularisation rate.

Methodologically, each hyperparameter was varied and the performance of the network for each combination was observed.
We chose to test the number of epochs with four values: 100, 500, 1000 and 1500.  
The considered learning rate was: 0.001, 0.005, 0.01, 0.05 and 0.1. 
The number of hidden layers was varied from 1 to 5, with the number of neurons per layer varying from 1 to 15. 
The weights were initialised as zero and also by using the initialiser HE \cite{he2015delving}.
\emph{Softsign} and \emph{hyperbolic tangent} functions were tested to activate hidden and cell states, while 
\emph{sigmoid} and \emph{hard-sigmoid} were tested as gates activation functions. 
Finally, the tested regularisation rate was of: 0.001, 0.003, 0.005 and 0.01.
}

After checking the performance of combinations of hyper-parameters, they are set to: 
$1000$ epochs; initial learning rate of $0.005$; and attenuation rate of $0.95$ every $300$ epochs.
The network was designed with 3 hidden layers, with 10 neurons each, and an output layer fully connected with 3 neurons and activation policy based on the method \emph{reLU} \cite{goodfellow2016deep}.
To the layers weights, we use the \emph{HE} initialisation model, while \emph{softsign} function is adopted to activate the hidden states and cell state; and \emph{sigmoid} function to activate the gates (see \Sect~\ref{subsec:LSTM_review}). 
To avoid over fitting, we use a regularisation rate of $0.003$.

\subsubsection{Evaluation}
our approach is evaluated, as usual, by comparing the \LSTM estimations of the test dataset. 
From the test dataset, we measure the \emph{coefficient of determination} (known as \rtwo), which is an acceptable measure for accuracy in regression. 
Tendency to $1$ indicates alignment of the model to the test dataset, while values tending to $0$ suggest otherwise. 

\Fig~\ref{fig:validLSTM} graphically illustrates the analysis for all \LSTM models ($\wm_{\week=1}$ to $\wm_{\week=6}$). 
The abscissa axis corresponds to the real output used in the network test ($\out[t]{Real}'$), which is normalised between 0 and 1 (see \Sect~\ref{data_pre_processing}), while the ordinate axis shows the estimated output, also normalised $\out[t]{Estimated}'$.
In general, the resulting graphs are linear, which indicates that \rtwo is close to 1.

\def\scaleIMG{0.158}
\begin{figure*}[t]
\psfrag{Axisx}[c][c]{\scriptsize{$ \\ \out[t]_{Real}'$}}
\psfrag{Axisy}[c][c]{\scriptsize{$\out[t]_{Estimate}'$}}
    \centering
	    \subfigure[][{\rtwo for $\wm_{\week=1}$}]{

		\includegraphics[scale = \scaleIMG]{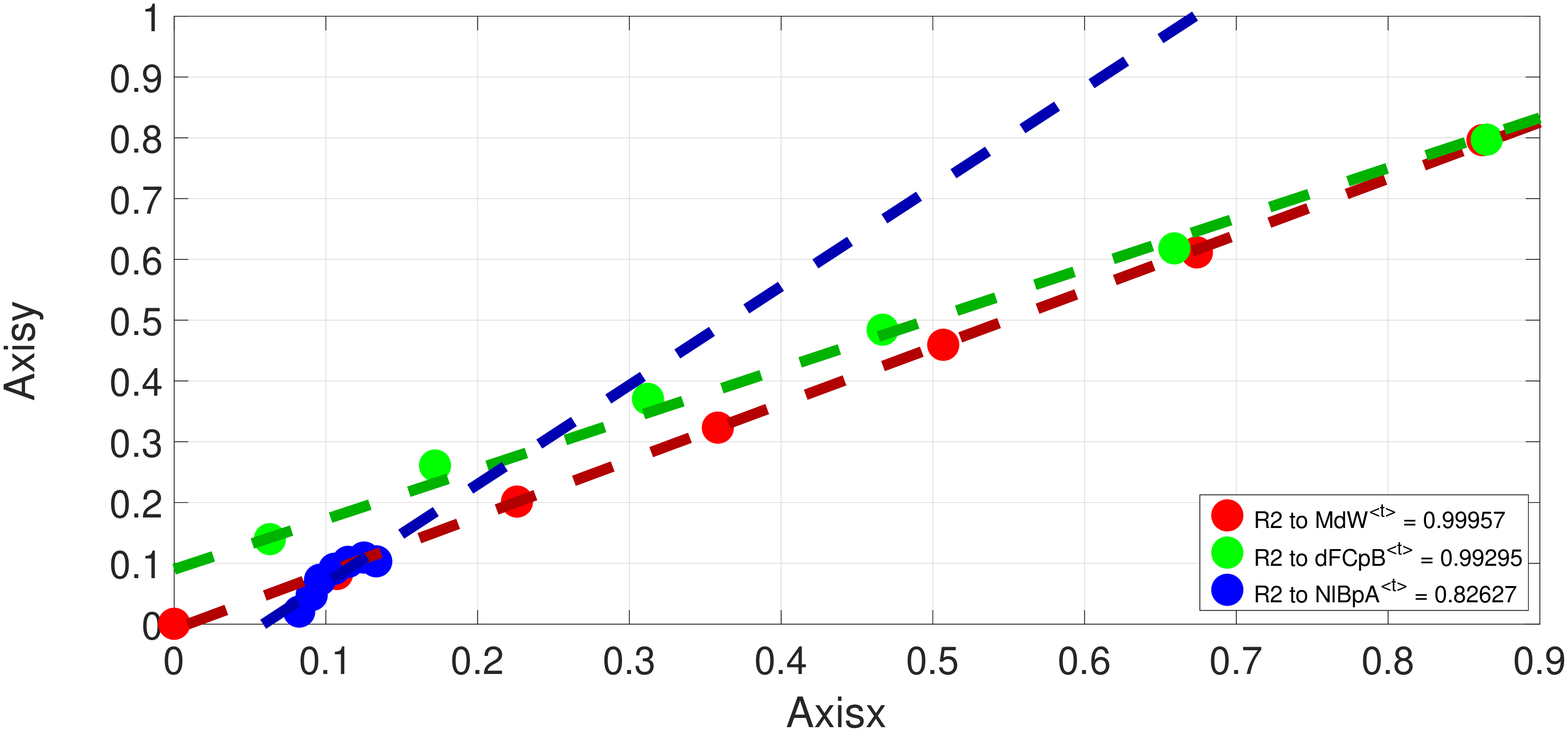}
		\label{fig:r2_w1}
	}
		\subfigure[][{\rtwo for $\wm_{\week=2}$}]{
		\includegraphics[scale = \scaleIMG]{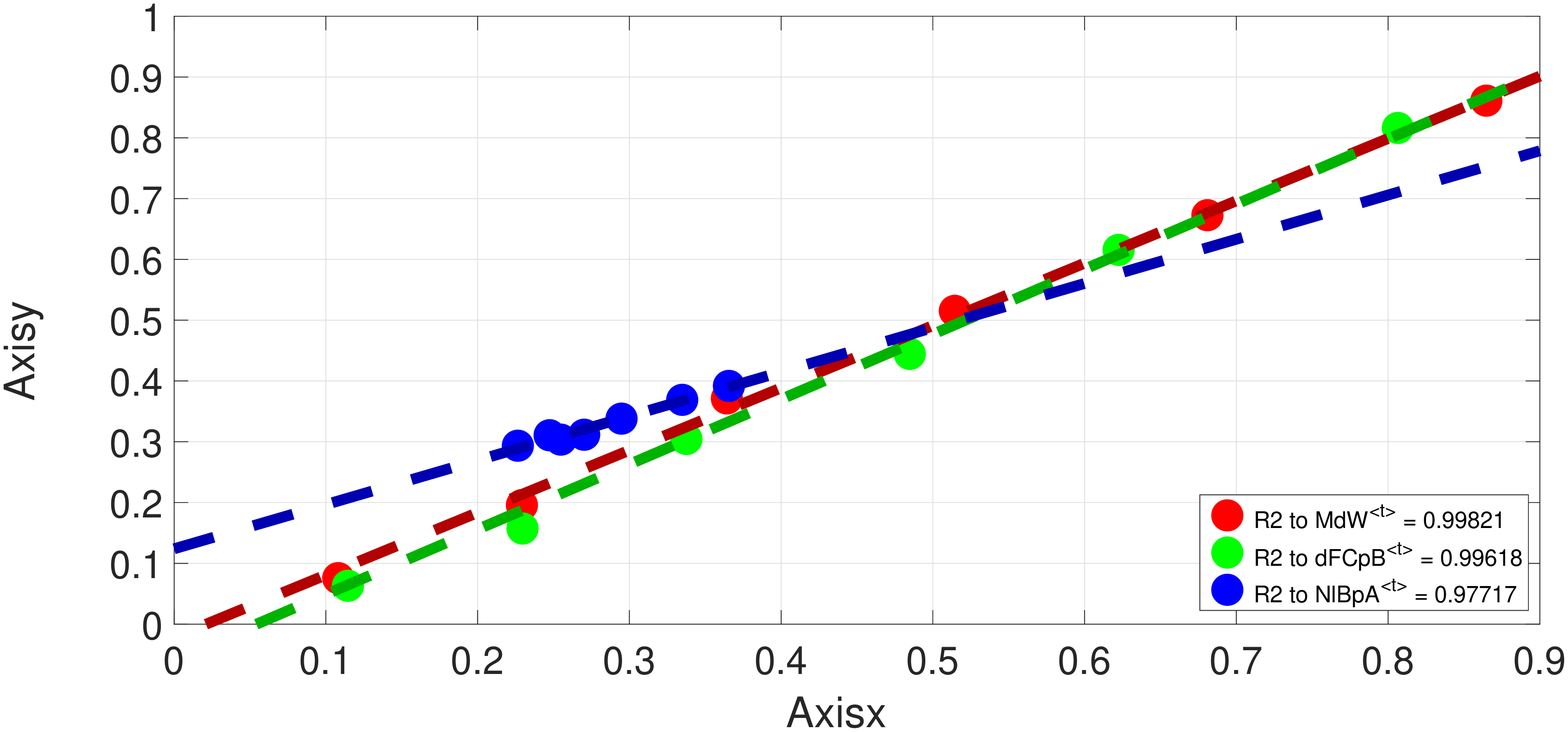}
		\label{fig:r2_w2}
	}
	    \subfigure[][{\rtwo for $\wm_{\week=3}$}]{
		\includegraphics[scale = \scaleIMG]{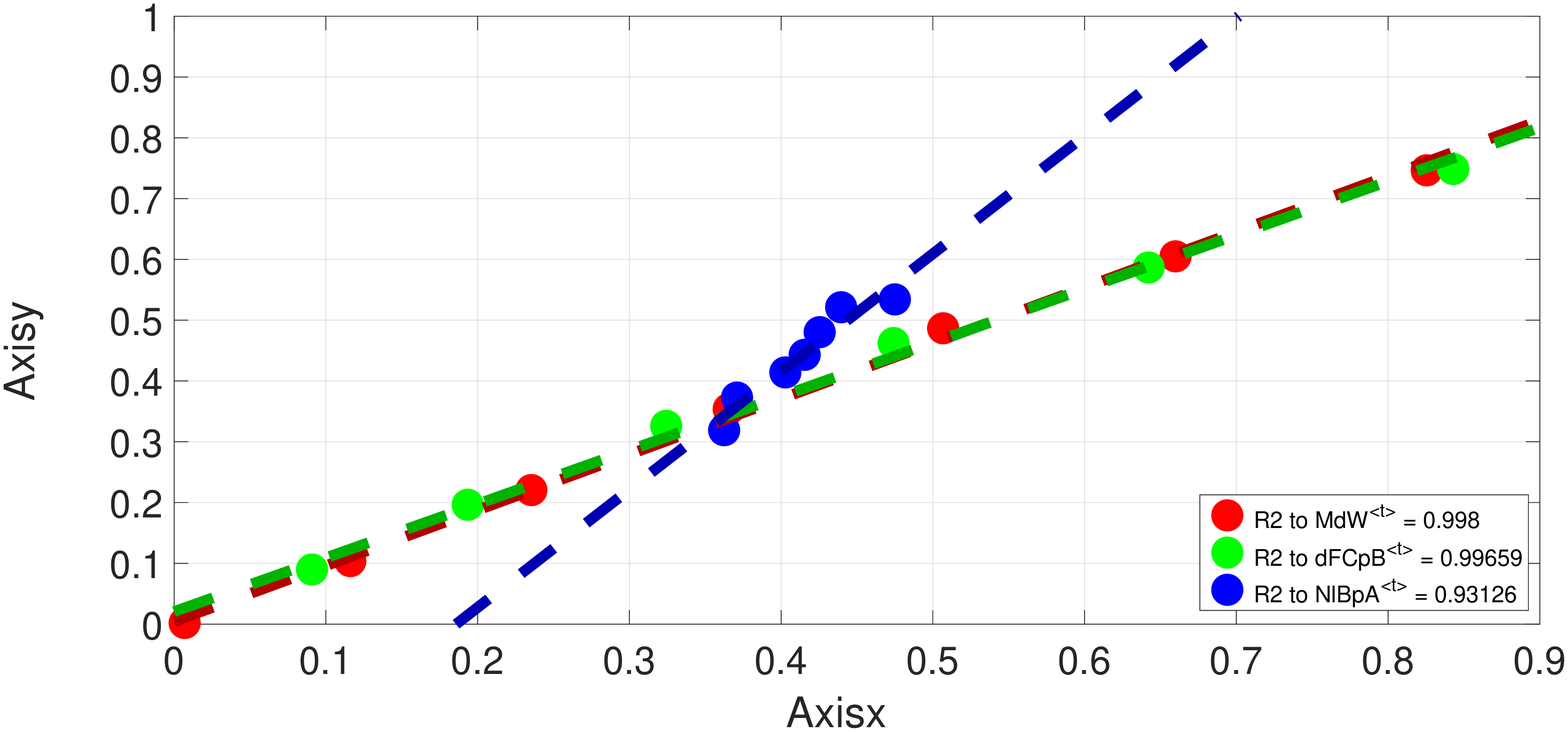}
		\label{fig:r2_w3}
	}
		\subfigure[][{\rtwo for $\wm_{\week=4}$}]{
		\includegraphics[scale = \scaleIMG]{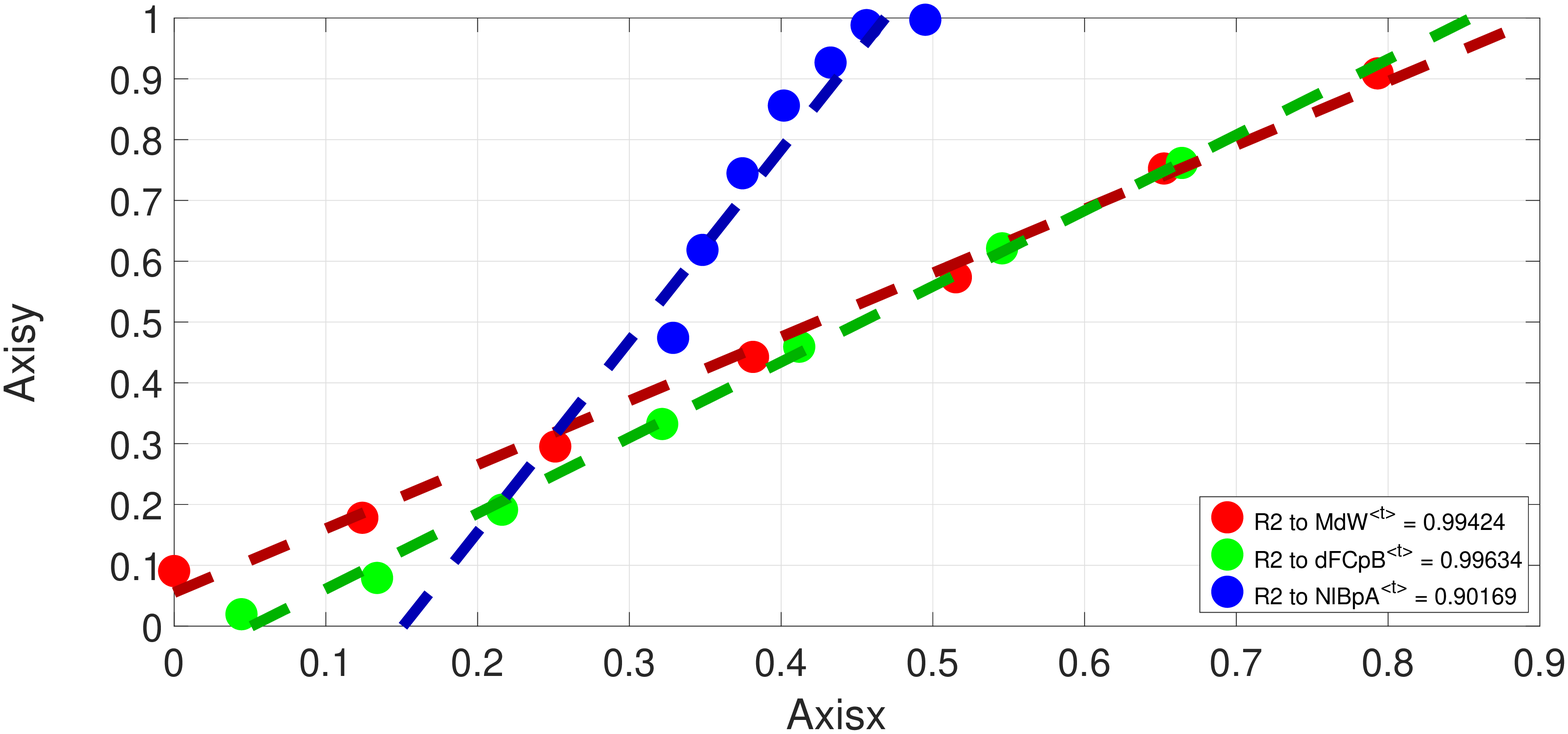}
		\label{fig:r2_w4}
	}
		\subfigure[][{\rtwo for $\wm_{\week=5}$}]{
		\includegraphics[scale = \scaleIMG]{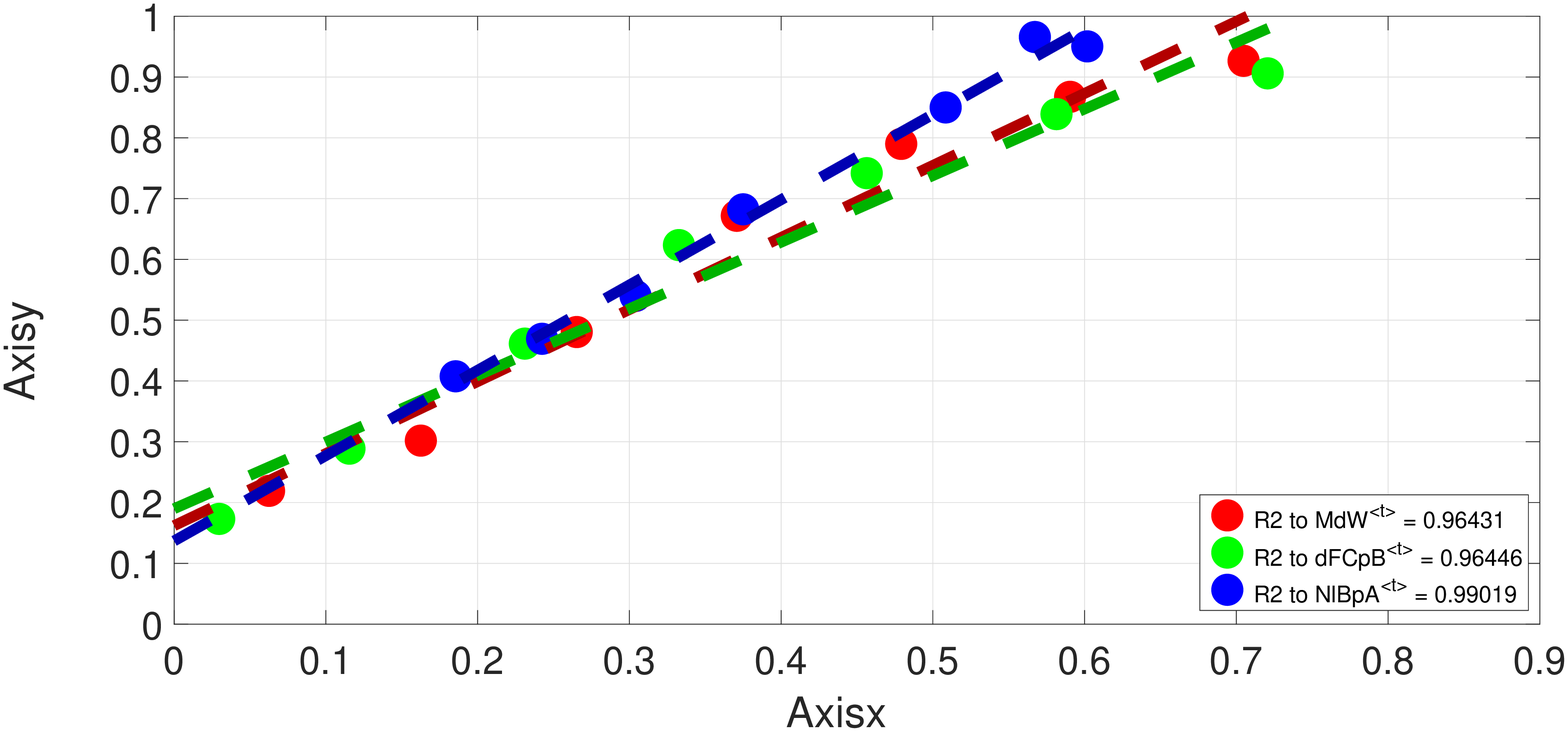}
		\label{fig:r2_w5}
	}
		\subfigure[][{\rtwo for $\wm_{\week=6}$}]{
		\includegraphics[scale = \scaleIMG]{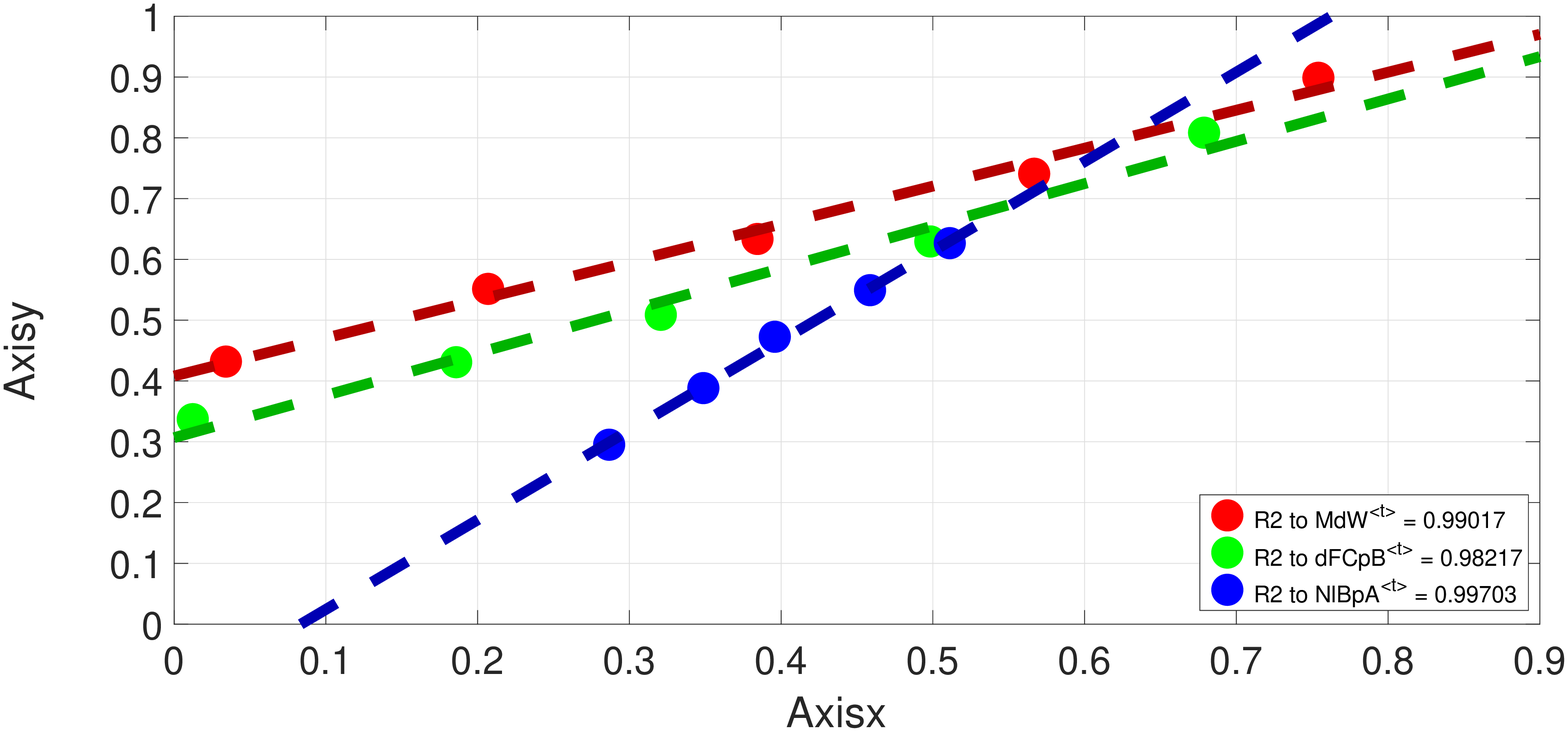}
		\label{fig:r2_w6}
	}
\caption{Coefficient of determination for the network output variables.} 
 \label{fig:validLSTM}
\end{figure*}

From \Fig~\ref{fig:validLSTM}, we see that the coefficients \rtwo for \MdW[t] and \dFCpB[t] are very similar for all six models. 
The weaker evaluation is found in $\wm_{\week=5}$ (\Fig~\ref{fig:r2_w5}), where \rtwo has an index of $0.964$. 
Remark that \rtwo close to 1 suggests a very reliable estimation for the first and second outputs of the \LSTM network.

The third output, i.e., \NlBpA[t] is the one that faces the major instability in comparison with the others, particularly when modelling the first week of production (\Fig~\ref{fig:r2_w1}). 
This follows \Fig~\ref{fig:mortes}, which shows an unstable deviation around the average for the initial period, which we believe is caused by manual interventions of the specialist over the flock, eliminating birds that, according to his judgement, are not healthy enough. 
As this action is completely random, the model faces difficulties to follow the specialist reasoning, and the result is a \rtwo index of $0.826$. 
For the remaining weeks, the \rtwo index fluctuates reasonably better, between $[0.90,0.99]$.

In comparison with the DNN approach in \cite{johansen2017data}, our \LSTM model performs significantly better. 
By inspecting the predictions \MdW[t] and \dFCpB[t], one finds a lack of accuracy of 3.42\% and 4.5\%. In comparison, the equivalent rate in \cite{johansen2017data} is on the order of 9\% and 6\%, for the same variables. \painter{As for the \ANN proposed in \cite{abreu2020artificial}, it returns \rtwo of $0.7918$ when predicting the \FCR[t], therefore also inferior to the \rtwo resulting from our \LSTM method.} 

From this point forward, we consider the \LSTM model validated, so that its 3 outputs can be used to estimate the \FCR[40] performance (by \Eq~\eqref{eq3}) under different action plans.
In order to systematise and simplify the call for the \LSTM execution, we denoted it by $\phi(\inpnor[t], \poutnor[t], \wm_{\week})$, and its return by $\outnor[t]$. 
Variable $\wm_{\week}$ identifies the \LSTM model $\wm$ for the week $\week$ of interest, while $\inpnor[t]$, $\poutnor[t]$, and $\outnor[t]$ are respectively normalised versions of $\inp[t]$, $\pout[t]$, and $\out[t]$. 

In this paper, as we aim to use the \LSTM periodically, we propose a methodology to execute it systematically through the six-weeks models under evaluation. Algorithm~\ref{alg:LSTM_implementation} implements the methodology for a one-week period. Later, in \Sect~\ref{subsec:integration}, Algorithm~\ref{alg:alggenetic_adaptive} iterates through Algorithm~\ref{alg:LSTM_implementation} to cover the whole six weeks.  



\subsubsection{\LSTM Implementation}
the proposed \LSTM-based architecture can be implemented as in Algorithm~\ref{alg:LSTM_implementation}. 
A \emph{Matlab}\textsuperscript{\textregistered}-compliant version of it can be found in \cite{Imp2020}. 

\begin{algorithm}[h]
\SetAlgoLined
\textbf{Procedure}: $\algtwo (\wm, \Vet[\maxi], \Vet[\mini], \Vet[\te])$\; 
\textbf{Input}: $\wm$, $\Vet[\maxi]$, $\Vet[\mini]$, $\Vet[\te]$\; \label{l:inp0}
\textbf{Output}: $\out[\week_S]$\;

\Begin{
    $\Vet[\tenor]=\Norm(\Vet[\te],\: \Vet[\maxi],\: \Vet[\mini])$\;  \label{l:normalize}
    \For{t=1 to 1}{
    $\inpnor[t]=\Vet[\tenor][1,7]^{T}$\; \label{l:inp1}
    $\poutnor[t]=\Vet[\tenor][8,10]^{T}$\; \label{l:inp3}
    $\outnor[t]=\phi(\inpnor[t],\poutnor[t], \wm)$\;\label{l:predit1}
    \emph{Update the \LSTM state}\; \label{l:update1}
}
    \For{t=2 to $\week_S$}{ \label{l:for_alg1}
    $\inpnor[t]=\Vet[\tenor][7t-3, 7t+3]^{T}$\; \label{l:inp2} 
    $\outnor[t]=\phi(\inpnor[t],\poutnor[t], \wm)$\; \label{l:predit2}
    \emph{Update the \LSTM state}\; \label{l:update2}
    }
    $return\; (\out[\week_S] = \DNorm(\Norm(\outnor[\week_S], \Vet[\maxi], \Vet[\mini])));$ \label{l:denormalize1}
}
 \caption{\LSTM implementation.}
 \label{alg:LSTM_implementation}
\end{algorithm}

Algorithm~\ref{alg:LSTM_implementation} receives four inputs required to execute the \LSTM model: 
a previously trained \LSTM model ($\wm$); 
vectors $\Vet[\maxi]$ and $\Vet[\mini]$ for (de)normalisation; and 
a vector carrying the weekly ($t$) input to the \LSTM (line \ref{l:inp0}), denoted $\Vet[\te]^{t}$, which has the following form: 
\begin{equation} \label{eq:inp_LSTM}
\Vet[\te]^{\left \langle t=1,2,\cdots,\week S \right \rangle} =[\inp[1] {^{\frown}} \pout[1] {^{\frown}} \inp[2] {^{\frown}} \cdots {^{\frown}} \inp[\week_S]].
\end{equation}
Remark that $\pout[t]$ has to be provided as input for the first day only, as the others are provided recursively for $t = 2,\cdots,\week_S$.

After $\Vet[\te]$ is normalised by \Eq~\eqref{eq6} (line \ref{l:normalize}), the inputs for the first day (variables $\inp[1]$ and $\pout[1]$) are selected from the 10 first locations of vector $\Vet[\tenor]$ ([1,7] for \inp[1] and [8,10] for \pout[1]). 
Then, a preliminary prediction is obtained for the first day (line \ref{l:predit1}) and the \LSTM state is updated (i.e., update $\Mat[C]^{t}$ as seen in \Fig~\ref{fig:lstm}) (lines \ref{l:update1} and \ref{l:update2}). 

Similarly, the inputs $\inp[t]$, for $t = 2,\cdots,\week_S$, are progressively selected from the respective positions of $\Vet[\tenor]$ (line \ref{l:inp2}). Remark that $\pout[t]$ is no longer necessary for $t = 2,\cdots,\week_S$, as it automatically follow from the output $\out[t]$ of the previous day $\dia-1$. Finally, a new prediction is made (line \ref{l:predit2}). 

Upon termination, a denormalised version of the \LSTM output for the last day of the week, i.e., $\out[\week_S]$, is returned (line \ref{l:denormalize1}).
It means the network prediction for the input $\Vet[\te]$. 
By repeating progressively the Algorithm~\ref{alg:LSTM_implementation} for $\wm_1,\cdots,\wm_6$, one can estimate the \FCR[40] to the whole flock. 
As there are 6 \LSTM models, \FCR[40] is then equivalent to the last day of $\wm_6$, i.e., $\FCR[40]=\FCR[\week_S]_{\week=6}$.

\subsubsection{A combinatorial problem appearing} \label{subsubsec:problem}
although the proposed \LSTM model is capable of estimating, with reasonable accuracy, the \FCR[40] for a given input action plan, such an estimation is individual and it has to be processed for each input. 
In this way, the \LSTM by itself cannot optimise the output, i.e., discover combinations of input variables that minimise the \FCR[40]. This has to be done manually, by repeating the Algorithm~\ref{alg:LSTM_implementation} for a number exhaustive of combinations. 

{\painterX
We actually implemented the \emph{exhaustive search} in order to illustrate this complexity. 
By constructing all possible combinations of variables values that compose an action plan, we fed the \LSTM model with each combination and observed the resulting \FCR[40]. The search space was delimited by the same maximum and minimum conditions observed in the real dataset. 
The temperature was varied in steps of 0.5 \graus, while humidity in steps of 1\% for each new test.  
This results in approximately $7.6\: e+185$ combinations, which can be computed in about 40 days of intensive processing. 
We consider this unfeasible and not escalable, and therefore a fancier alternative is presented next, based on \GAs.
}

\subsection{Integrating \GA and \LSTM models} \label{subsec:integration}

From this section forward, the \LSTM model proposed in \Sect~\ref{subsec:lstm} is taken as a fitness function for a \GA to be extended from the Algorithm~\ref{alg:alggenetic}. This combination assigns computational efficiency to the search method (\wrt exhaustive approaches), at the same time as the \LSTM memory is exploited. The price to be paid for this combination is that the globally optimal action plan is eventually not found, and replaced by a reasonably close approximation of it.

\begin{figure*}[t]
\psfrag{S}[c][c]{\scriptsize{Y}}
\psfrag{N}[c][c]{\scriptsize{N}}
\psfrag{wm6}[c][c]{\scriptsize{$\wm_{\week=6}$}}
\psfrag{gs1}[c][c]{\scriptsize{$\Gen[S]=38$}}
\psfrag{Popinitial}[c][c]{\scriptsize{Initial Population}}
\psfrag{LSTM}[c][c]{\scriptsize{\LSTM}}
\psfrag{Ps}[c][c]{\scriptsize{$\Pop[S] = 200$}}
\psfrag{Restricoes}[c][c]{\scriptsize{Restrictions}}
\psfrag{Norm}[c][c]{\scriptsize{Normalization }}
\psfrag{Predict}[c][c]{\scriptsize{Prediction}}
\psfrag{Denorm}[c][c]{\scriptsize{Denormalization}}
\psfrag{Fitness}[c][c]{\scriptsize{Fitness}}
\psfrag{comp1}[c][c]{\scriptsize{$\wm_{\week=6}?$}}
\psfrag{ff1}[c][c]{\scriptsize{$\ff=\FCR[40]$}}
\psfrag{ff2}[c][c]{\scriptsize{$\ff= \mean \left | \out[\week_S]_{\week}-\pout[1]_{\week+1} \right |$}}
\psfrag{Ind}[c][c]{\scriptsize{Individuals}}
\psfrag{Avaliable}[c][c]{\scriptsize{evaluated ?}}
\psfrag{Selection}[c][c]{\scriptsize{Selection}}
\psfrag{father}[c][c]{\scriptsize{$\Vet[p_1],\Vet[p_2]=\SU(\ff)$}}
\psfrag{Crossolver}[c][c]{\scriptsize{Crossover}}
\psfrag{heurist}[c][c]{\scriptsize{\Vet[child]=$\HC(\Vet[p_1],\Vet[p_2],\beta)$}}
\psfrag{Mutation}[c][c]{\scriptsize{Mutation}}
\psfrag{Stop}[c][c]{\scriptsize{Stop Criteria}}
\psfrag{Atuali}[c][c]{\scriptsize{Load Week Model}}
\psfrag{Adaptive}[c][c]{\scriptsize{$\Vet[\Gen]=\AM(\Vet[child])$}}
\psfrag{Popul}[c][c]{\scriptsize{Complete}}
\psfrag{complete}[c][c]{\scriptsize{population?}}
\psfrag{Atendido}[c][c]{\scriptsize{Achieved?}}
\psfrag{week}[c][c]{\scriptsize{Full}}
\psfrag{compl}[c][c]{\scriptsize{Weeks?}}
\psfrag{ic}[c][c]{\scriptsize{\Vet[i\_c]}}
\psfrag{FAP}[c][c]{\scriptsize{\Mat[\fap]}}
\psfrag{wm}[c][c]{\scriptsize{$\wm_{\week-1}$}}
\psfrag{gs2}[c][c]{\scriptsize{$\Gen[S]=52$}}
    \centering
	\includegraphics[width=15cm]{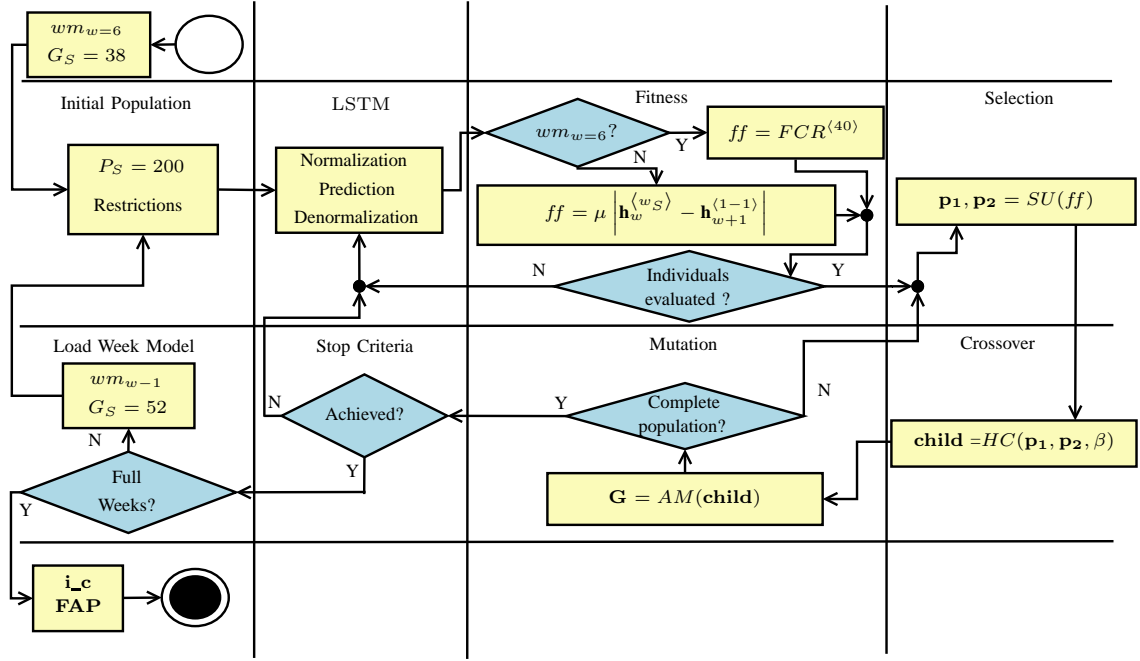} 
	\caption{Ray diagram illustrating the \LSTM and \GA integration.}
\label{fig:raia_diagram}
\end{figure*}

\painter{
Algorithm~\ref{alg:alggenetic_adaptive} integrates \LSTM and \GA. 
A \emph{Matlab}-compliant implementation of the algorithm can be found in \cite{Imp2020}, and the reasoning behind it is illustrated in \Fig~\ref{fig:raia_diagram}. 
}

\begin{algorithm}[h]
\SetAlgoLined
\textbf{Input}: $\wm_{\week}$, $\Vet[\maxi]_{\week}$, $\Vet[\mini]_{\week},~\week=1,\cdots,6$\; 
\textbf{Output}: $\Mat[\fap]$, $\Vet[i\_c]$\; 

\Begin{
$\Gen[S]_{\week=6}=38$\; \label{l:set1}
$\Gen[S]_{\week=5,4,\cdots,1}=52$\;\label{l:set2}
set $\Pop[S] = 200$, $\beta=0.6$\; \label{l:definition1}

   \For{\week = 6 to 1}{  \label{for}
 
    set $\Mat[A] = [2\:\Gen[S]_{\week}$ x $\Gen[S]_{\week}]$\; \label{l:restriction:1}
    set $\Vet[x] = [\Gen[S]_{\week}$ x $1]$\;\label{l:restriction:2}
    set $\Vet[b] = [2\:\Gen[S]_{\week}$ x $1]$\;\label{l:restriction:3}
    
    set $\Mat[\pop] = [\Pop[S]$ x $\Gen[S]_{\week}]$\;\label{l:population:2}
    $\Mat[\pop]=\Pop[I](\Mat[A],\Vet[x],\Vet[b])$\; \label{l:population:3}
 \While{Some stopping criterion is not met}{ \label{while}

 \For{i=1 to $\Pop[S]$}{\label{l:loopfit}
     $\out[\week_S]_{\week} = \algtwo (\wm_{\week}, \Vet[\maxi]_{\week}, \Vet[\mini]_{\week}, \Mat[\pop](i,\: :)^{T} $\; \label{l:fitness:1} 
\Ddot{}
    \eIf{\week == 6}{
   $\ff = \FCR[\week_S]_{\week}$\; \label{l:fitness:5}
   }{
  $\ff= \mean \left | \out[\week_S]_{\week}-\pout[1]_{\week+1} \right |$ \label{l:fitness:6}

   }
}
  \For{j=1 to $\Pop[S]$}{\label{l:loopNewPop}
     $\Vet[p_1],\Vet[p_2] = \SU(ff)$\; \label{l:selection:1}
     $\Vet[child]=\HC(\Vet[p_1],\;\Vet[p_2],\beta\;);$\\  \label{l:crosolver:1}
     $\Vet[\Gen]=\AM(\Vet[child])$\; \label{l:mutation:1}
     $\Mat[P](j,\: :)=\Vet[\Gen]$\; \label{l:new_pop}
   }  
    
 } \label{whileend}
  $ \pout[1]_{\week}= \Vet[\Gen](8:10)$\; \label{l:atribuition}
 $\Mat[\fap](\week,\: :) = \Vet[\Gen];$ \label{l:paweek}
 
 }
 $\Vet[i\_c]\: =\: \pout[1]_{\week}$\; \label{2l:recomendation} 
 $return\: (\Vet[i\_c],\: \Mat[\fap]);$ \label{l:return_plain} 
}
 \caption{GA and \LSTM Network Integration}
 \label{alg:alggenetic_adaptive}
\end{algorithm}

The algorithm receives the six models of the \LSTM neural network ($\wm_{\week=1}$ to $\wm_{\week=6}$) and the vectors $\Vet[\maxi]_{\week}$ and $\Vet[\mini]_{\week}$ for (de)normalisation of the six models. Later, these \painter{information} will be entered for the \LSTM function call.

Then, lines \ref{l:set1} and \ref{l:set2} set the size of vector $\Vet[\Gen]$ (defined by $\Gen[S]$, see \Sect~\ref{subsub:populationandrestrictions}) for each week (\week).
In practice, the vector $\Vet[\Gen]$ is a weekly candidate action plan (in addition, it also contains $\pout[t]$ for the first day, identical to the vector $\Vet[\te]$ presented in \Eq~\eqref{eq:inp_LSTM}, i.e. $\Vet[\Gen] = \Vet[\te]$. Therefore, $\Gen[S]$ is calculated by $7\week_S + 3$, that is, 7 inputs $\inp[t]\: \times \: \week_S$ days plus 3 inputs of $\pout[t]$.

Line \ref{l:definition1} quantifies the population size ($\Pop[S]$) and $\beta$. $\Pop[S]=200$ (see \Sect~\ref{subsub:populationandrestrictions} ) and $\beta=0.6$ (see \Sect~\ref{subsubcross}).
 
Lines \ref{l:restriction:1}, \ref{l:restriction:2}, and \ref{l:restriction:3} define the structure of restrictions (see \Eq~\eqref{eq5}). This is associated with the search space of the \GA, limited between $\Vet[\mini]$ and $\Vet[\maxi]$.

Line \ref{l:population:2} sets the dimension of the matrix $\Mat[\pop]$. In practice, each line of $\Mat[\pop]$ carries a vector $\Vet[\Gen]$. Therefore, $\Pop[S]=200$ implies 200 candidate action plans for each $\week$ to be evaluated by the fitness function (that is, the Algorithm~\ref{alg:LSTM_implementation}).
Line \ref{l:population:3} constructs the initial population $\Pop[I]$ randomly (but limited the same restrictions described previously) and assigns it to the matrix $\Mat[\pop]$. 

By entering the loop on line \ref{l:loopfit}, the algorithm applies two fitness functions (\ff), both defined over the return $\out[t]$ from the Algorithm~\ref{alg:LSTM_implementation}. 
The first, on line \ref{l:fitness:5}, is exclusive for the sixth week ($\week=6$) and it aims to minimise the $\FCR[\week_S]_{\week=6}$ (\Eq~\eqref{eq3}). 
The second, on line \ref{l:fitness:6}, applies for the other weeks ($\week=5,4,\cdots,1$) and it aims to minimise (in terms of mean and absolute values) the difference between $\out[{\week_S}]$, estimated by model $\wm_{\week}$ (i.e., $\out[{\week_S}]_{\week}$), and $\pout[1]$ of the model $\wm_{\week+1}$ (i.e., $\pout[1]_{\week+1}$), which has already been evaluated (since the execution recurs from $\wm_6$ to $\wm_1$ as in line \ref{for}).

Thus, with the \ff presented in line \ref{l:fitness:6} we guarantee that the conditions $\pout[1]$, required as input for the $\wm_{\week+1}$ model, are meet by $\out[{\week_S}]$ of model $\wm_{\week}$.
Upon evaluation of \ff, the action plans enter the loop on line \ref{l:loopNewPop} and are selected using the \SU method on line \ref{l:selection:1} (see justification in \Sect~\ref{subsub:fitnes}).
\SU receives the fitness function \ff as input and returns the $\Vet[p1]$ and $\Vet[p2]$, i.e., the two most suitable action plans. 

Then, $\Vet[p1]$ and $\Vet[p2]$ exchange their genes $\gen[i]$, and generate an improved action plan ($\Vet[child]$) by the \emph{crossover method} (\HC) (line \ref{l:crosolver:1}, see \Eq~\eqref{eq4}). 

Finally, the \emph{Adaptive Mutation} (\AM) method (line \ref{l:mutation:1}) was used to multiply some $\gen[i]$ of the new action plan ($\Vet[child]$) by a (very low) random value, still under the restrictions (see \Eq~\eqref{eq5}). 
\AM receives $\Vet[child]$ and returns $\Vet[\Gen]$, a new candidate action plan (line \ref{l:mutation:1}). 
Then, \emph{selection} (line \ref{l:selection:1}), \emph{crossover} (line \ref{l:crosolver:1}) and \emph{mutation} (line \ref{l:mutation:1}) steps are repeated (line \ref{l:loopNewPop}) until a new population is completed, possibly more adapted than the previous one (line \ref{l:new_pop}).

The entire process described so far (line \ref{while} through line \ref{whileend}) is repeated until any of the stopping criteria presented in \Sect~\ref{subsubsec:stopcrite} is reached. 
When this happens, the values of $\pout[1]_{\week}$ are updated for use in the next step (line \ref{l:atribuition}, note from \Eq~\eqref{eq:inp_LSTM} that its locations are from 8 to 10) and $\Vet[\Gen]$ is assigned to the matrix final action plan ($\Mat[\fap]$ in line \ref{l:paweek}).

Algorithm~\ref{alg:alggenetic_adaptive} is repeated from week 6 to week 1, i.e., $\week=6,5,\cdots,1$ (line \ref{for}).
By the end $\week=1$, the inputs $\pout[1]_{1}$ carry the values for \MdW[t], \dFCpB[t], and \NlBpA[t] (see Table~\ref{tab:variable}) of day 0, i.e., the initial conditions ($\Vet[i\_c]$) for the arrival of the birds at the \house (line \ref{2l:recomendation}).
The algorithm then returns $\Vet[i\_c]$ and $\Mat[\fap]$ with the action plan for 40 days (line \ref{l:return_plain}).

\subsubsection{Proposed integration} \label{subsub:expoproposed}
\begin{figure}[!htb]
    \centering
	\psfrag{Axisy}[c][c]{\scriptsize{Feed Consumption Rate}}
	\psfrag{Axisx}[c][c]{\scriptsize{Iterations}}
	 \includegraphics[width=8cm]{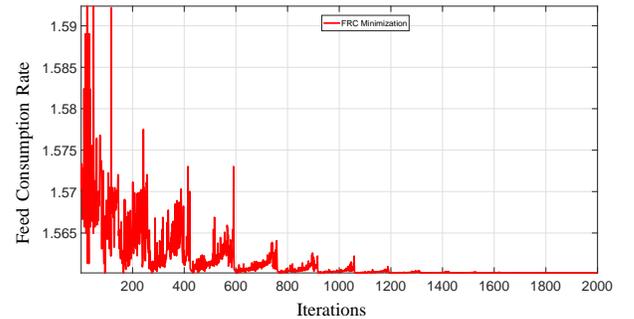}
     \caption{\FCR[40] improvement process.}
    \label{fig:ga_week6}
\end{figure}

we describe in \Fig~\ref{fig:ga_week6} a step-by-step to improve the \FCR[40] along the iterations.

After $1400$ iterations the proposed algorithms start to stabilise.
This leads search to stop by the timeout criterion with no changes in the fitness function.
The \FCR[40] upon stabilisation is $1.5602$. 
However, this value is exclusively linked to $\wm_{6}$, i.e., as the Algorithm~\ref{alg:alggenetic_adaptive} performs the search from $\wm_{6}$ to $\wm_{1}$, the value of $1.5602$ is guaranteed if and only if the progressive error between the difference of $\out[\week_S]_{\week}$ (estimated by the $\wm_{\week}$ model) and $\pout[1]_{\week+1}$ (of the model $\wm_{\week+1}$) is null (see Algorithm~\ref{alg:alggenetic_adaptive}, line \ref{l:fitness:6}). The following Table~\ref{tab:ga_other_weeks} shows step-by-step this errors:

\begin{table}[h]
\centering
\scriptsize
\begin{tabular}{c|ccc|ccc} \hline
\multirow{2}{*}{} & \multicolumn{6}{c}{$ |\out[\week_S]_{\week}-\pout[1]_{\week+1}|$}  \\ \cline{2-7} 
$w$ & \multicolumn{3}{c|}{\textbf{Error}} & \multicolumn{3}{c}{\textbf{Relative Error (\%)}} \\ \cline{2-7}

& \scriptsize{\MdW[t]}          
& \scriptsize{\dFCpB[t]}         
& \scriptsize{\NlBpA[t]}        
& \scriptsize{\MdW[t]}         
& \scriptsize{\dFCpB[t]}            
& \scriptsize{\NlBpA[t]}             \\ \hline
5                              & 17.4840          & 0.0066         & 0.1448        & 0.8335         & 0.2230            & 1.0269            \\ 
4                              & 32.7932          & 0.0011         & 0.0092        & 2.3440         & 0.1960            & 0.0639            \\ 
3                              & 15.7567          & 0.0070         & 0.1289         & 1.7667         &  0.7056            & 0.9059            \\ 
2                              & 0.0000          & 0.0080         & 0.0765         & 0         & 2.2396            & 0.5358            \\
1                              & 1.6142          & 0.0015         & 0.1461         & 1.0338         & 1.4681            & 1.0176            \\ \hline
\end{tabular}
\caption{Errors and relative erros comparison.}
\label{tab:ga_other_weeks}
\end{table}

Columns 2 and 5 show, respectively, the error and the relative error for the output \MdW[t]. 
This output has relative error values ranging from 0\% to 2.344\%, with the worst index observed at week 4.
The same analysis is performed for the output \dFCpB[t] (columns 3 and 6), for which the relative error varies from 0.1960\% to 2.2396\%. 
Columns 4 and 7 assess the output \NlBpA[t], for which the relative error varies between 0.0639\% and 1.0269\%. 


Observe that the errors rates are minor, i.e., the values $\out[\week_S]_{\week}$ of the current model ($\wm_{\week}$) are approximately the same as $\pout[1]_{\week+1}$ of the next model ($\wm_{\week+1}$). 
This suggests that the \emph{estimated} $\FCR[40] = 1.5602$ is accurate and it will not change much when the action plan is applied progressively. 
It remains to discover how much it changes, i.e., discover which is the \emph{resulting} \FCR[40] after an action plan is progressively applied (instead of recursively) through the six weeks. Minor variations are possible, but they are expected to be in accordance with Table~\ref{tab:ga_other_weeks}. 
To avoid terminology confusion, let us call by $\FCR[\est]= 1.5602$ the \emph{estimation} from Algorithm~\ref{alg:alggenetic_adaptive}, and by $\FCR[\res]$ the \emph{result} of the progressive application of the action plan. In this case, $\FCR[\res]$ reveal the real application of the estimated action plan by the \painter{specialist} over the process. 

We start by presenting in \Fig~\ref{fig:planactionfinal} the action plan $\Mat[\fap]$ returned by the Algorithm~\ref{alg:alggenetic_adaptive} after it iterates recurrently through the six weeks, leading to $\FCR[\est]= 1.5602$. 
The abscissa axis shows the input \dia of the \LSTM network (see \Fig\ref{fig:arquitetura_rede}). 
The ordinate axis shows the inputs \Min[T], \Avg[T], and \Max[T] (\Fig~\ref{fig:1temp}); and the inputs \Min[H], \Avg[H], and \Max[H] (\Fig~\ref{fig:2humi}) that were discovered by the \GA in order to reach $\FCR[\est]= 1.5602$. 

\def\scaleIMG{0.16}
\begin{figure}[!htb]
    \centering
    \psfrag{Axisx}[c][c]{\scriptsize{Day (\dia)}}
	    \subfigure[][{Temperature}]{
	    \psfrag{Axisy}[c][c]{\scriptsize{Temperature [\graus]}}
		\includegraphics[scale = \scaleIMG]{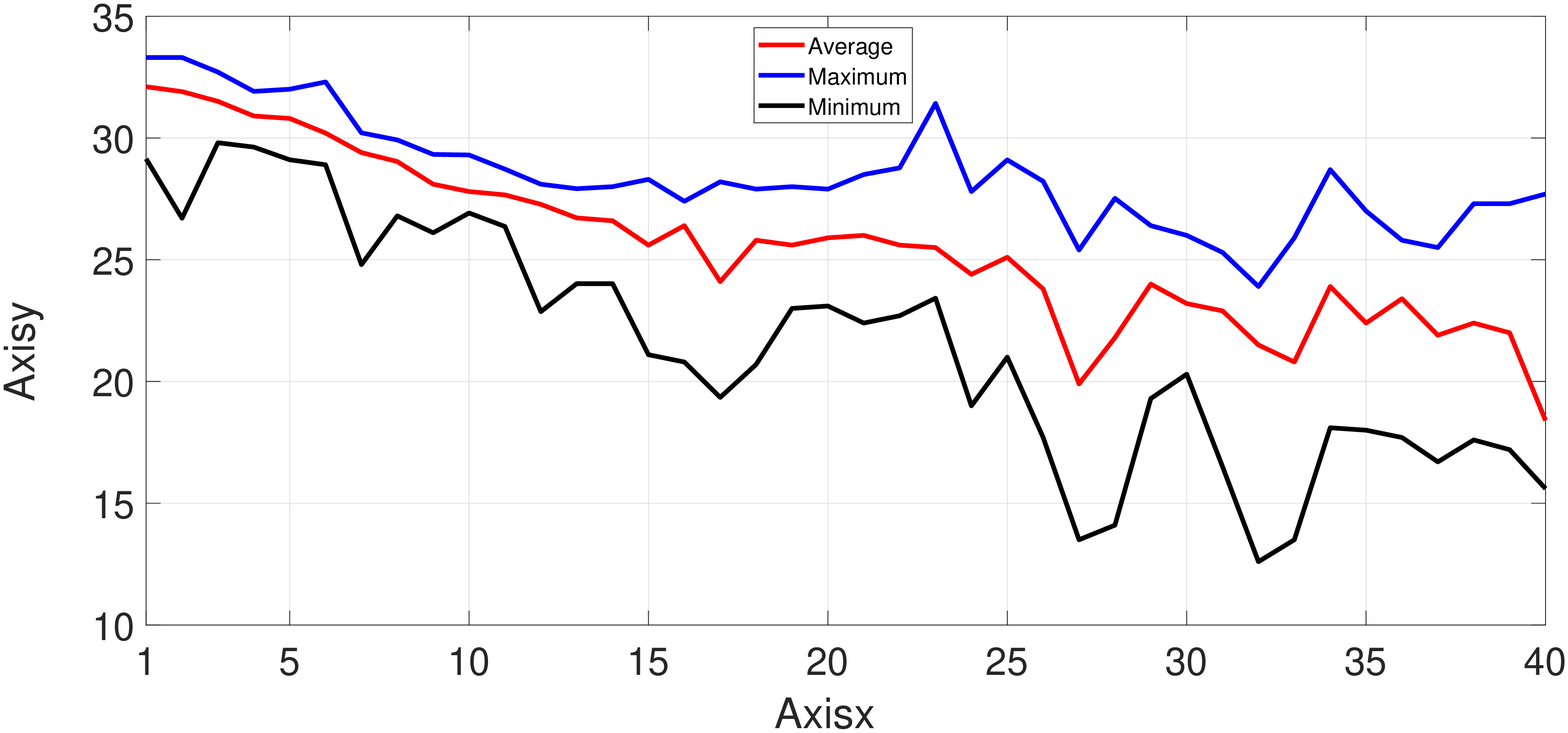}
		\label{fig:1temp}
	}
		\subfigure[][{Humidity}]{
	    \psfrag{Axisy}[c][c]{\scriptsize{Humidity [\%]}}
		\includegraphics[scale = \scaleIMG]{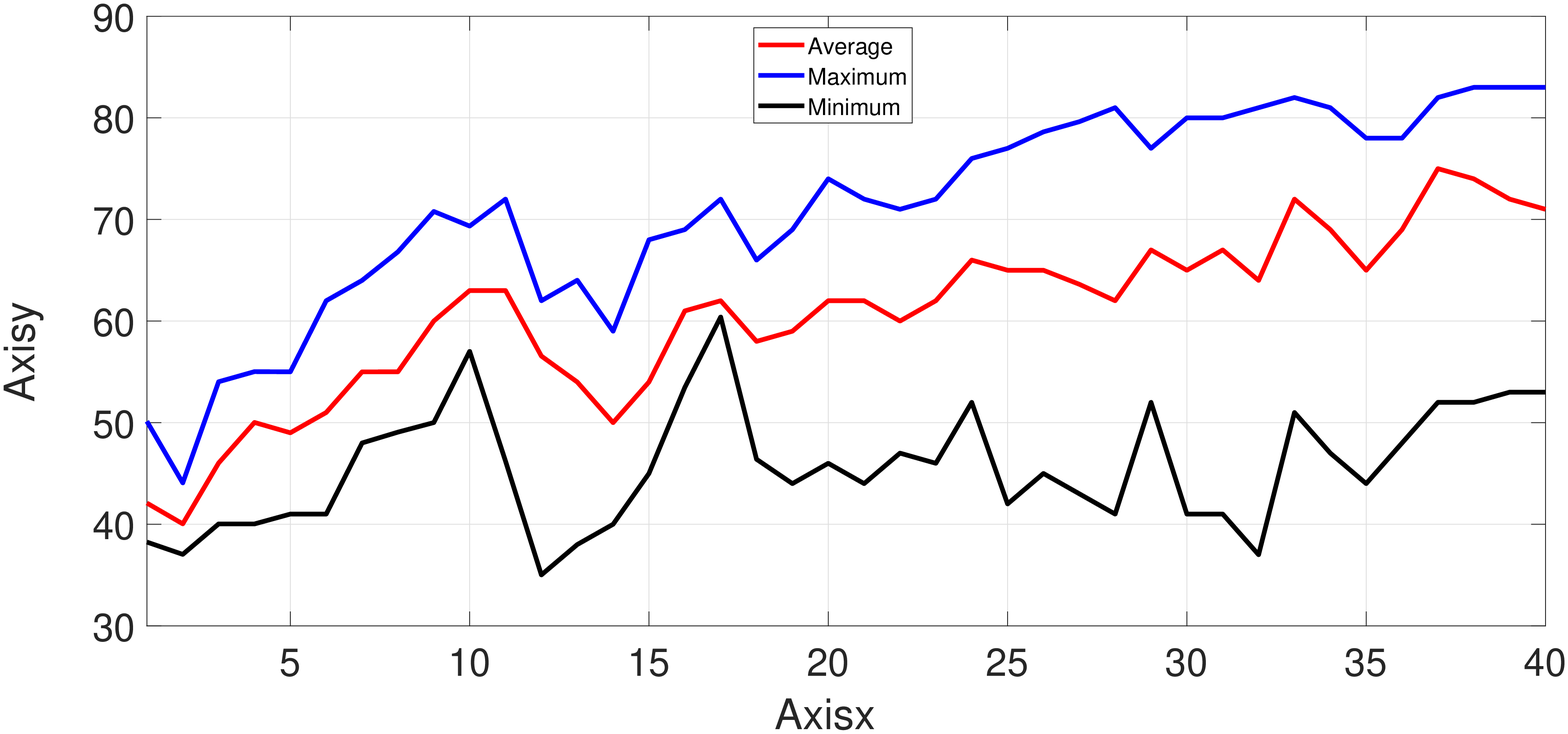}
		\label{fig:2humi}
	}
\caption{Action plan suggested by our approach.} 
 \label{fig:planactionfinal}
\end{figure}

The strategy suggested by the algorithm to reach the target $\FCR[\est]= 1.5602$ allows certain level of variations around the average, which is consistent with the control practice in \houses.
In addition to $\Mat[\fap]$, the algorithm~\ref{alg:alggenetic_adaptive} also returns the $\Vet[i\_c]$ for the birds before they arrive at the \house (i.e, for $\dia=0$, see lines \ref{2l:recomendation} and \ref{l:return_plain}). 
The calculation of $\Vet[i\_c]$ stems from the recursive nature of our algorithm, that is, as it starts from week 6 and recursively reaches week 1, then (see \Fig\ref{fig:arquitetura_rede}) the \painter{$\MdW[0]$, $\dFCpB[0]$, and $\NlBpA[0]$} carry the initial setup for the flock. 
For the evaluated dataset, it follows that: \painter{$\MdW[0] = 42.01\: \grama$; $\dFCpB[0] = 0\:\Kg$; and $\NlBpA[0] = 14.735\:bird/\mtwo$}. 
That is, broilers should arrive weighing $42.01\: \grama$, with null feed consumption, and in quantity of $14.735$ $bird/\mtwo$.

Remark that $\Vet[i\_c]$ setup impose two important assumptions for the efficiency of out method: number of birds per area (\NlBpA[0]) and initial weighing (\MdW[0]). Both have practical relevance, as \emph{in loco} visits evidence that frequent and divergent variations apply for the initial values for \MdW[0] and \NlBpA[0]. 
From now on, we assume them as well defined assumptions. 

Now, since we have estimated an action plan for each day of a flock ($\Mat[{\fap}]$), including the initial conditions ($\Vet[{i\_c}]$), we are in position to derive the obtained $\FCR[{\res}]$. The procedure consists in applying the Algorithm \ref{alg:LSTM_implementation}, this time in progressive order, i.e., from $\wm_{\week}=1$ to $\wm_{\week}=6$, having $\Mat[{\fap}]$ and $\Vet[{i\_c}]$ as input for each respective model. 
Upon termination of Algorithm \ref{alg:LSTM_implementation}, we capture the 3 outputs from the last day of $\wm_6$ (i.e., $\out[\week_S]_{\week=6}$) and apply \Eq\eqref{eq3} to obtain $\FCR[\res]$. It is expected a very close relationship between $\FCR[\res]$ and $\FCR[\est]$, considering the statistical correctness of Table~\ref{tab:ga_other_weeks}. 
For the particular dataset, $\FCR[\res] = 1.5610$ was found for the 40-day flock, which is only $0.0512\%$ higher than $\FCR[\est]= 1.5602$.

\subsubsection{Quantitative analysis} \label{secsec:validation}
{\painterX
in order to assess how representative our estimations are, we present in the following three comparisons. 
The first shows how far $\FCR[\res]$ is from the best performance possible to be obtained for a given broiler breed (denote $\FCR[\opt]$); 
the second reveals how good $\FCR[\res]$ is in comparison with a group of other candidate action plans; and 
the third tests $\FCR[\res]$ against synthetic action plans. 

\begin{itemize}
    \item 
According to \cite{rossperformace:Online}, the optimal performance \FCR[\opt] (the minimum value) for \emph{Ross}-breed broilers after 40 days is $1.558$. This involves genetic factors, in addition to conditions of feed and housing. 
Therefore, the performance of $\FCR[\res]=1.5610$, resulting from our method, suggests a very close approximation to the \glob (only $0.192\%$ greater than \FCR[\opt]). 
We credit this accuracy to the modular way the \LSTMs were implemented, splitting the broiler growth into 6 submodels, to the long-term memory of \LSTMs, and to the effectiveness of multi-objective \GA-based searches. 
    \item 
The second study compares the bio-inspired $\FCR[\res]$ with empiric performance (denote \FCR[\emp]), i.e., with action plans chosen by a specialist. 
Empiric action plans were collected from the real dataset, and the specialist who had the best performance was considered. 
This resulted in $\FCR[\emp] = 1.640$, therefore $0.0790$ (or $5\%$) greater than $\FCR[\res] = 1.5610$. 
Although the difference seems to be minor, the choice for the bio-inspired model implies a saving of $-0.0790$ \Kg (or $-79$ \grama) of feed per bird, for each $1$ \Kg of meat it produces. 
For an average slaughter weight of $2.8$ \Kg, each broiler requires $221.20$ \grama less feed to be produced. 
Considering a conventional \house with about $34000$ broilers, this leads to an amount of $7520,8$ \Kg of feed saving during a single flock.
    \item 
We further construct 1000 more synthetic action plans to simulate different specialists handling a flock. 
The same setup, data domains, and conditions were considered. 
By feeding the \LSTM with these action plans, we discover the performance of each synthetic specialist. 
The best possible performance found was $\FCR[\sint]= 1.5614$, therefore still worse than $\FCR[\res] = 1.5610$.
\end{itemize}

Regarding the computational cost, the Algorithm \ref{alg:alggenetic_adaptive} converged after 96 minutes processing on a setup  Intel\textsuperscript{\textregistered} Core\texttrademark i5, 2.20 GHz processor, with 6 GB of RAM. 
This points to an average of 16 minutes for processing each of the six, previously trained, \LSTM models. 
In addition, we believe that computational cost is not a problem, since Algorithm \ref{alg:alggenetic_adaptive} is only executed in times of readaptation.}



\section{A control technology solution} \label{sec:implement}

\painter{Now, we show how our proposal can be implemented, transferred, and combined with control technology in order to support optimised automatic control of \PCs.} 
A \PC as defined in \Fig~\ref{fig:Cond}, is the architecture that can be mostly benefited by our proposal, given its distributed and heterogeneous nature. Therefore, we use it to illustrate our results. 

\subsection{Implementation methodology}\label{subsec:meth}

\begin{figure}[!htb]
\psfrag{Poultry_House_1}[c][c]{\small{\PH[1]}}
\psfrag{Poultry_House_2}[c][c]{\small{\PH[2]}}
\psfrag{Poultry_House_n}[c][c]{\small{\PH[n]}}
\psfrag{Comunication}[c][c]{\tiny{Communication}}
\psfrag{Switch}{\tiny{switch}}
\psfrag{Aquisicao_Variables}[c][c]{\tiny{data acquisition}}
\psfrag{Optimized_Action_Plan}[c][c]{\tiny{ideal action plan}}
\psfrag{Comunication}[c][c]{\tiny{\textbf{Communication}}}
\psfrag{ModbusTCP}[c][c]{\tiny{Modbus RTU}}
\psfrag{SCADA}[c][c]{\footnotesize{\SCADA}}
\psfrag{Server_OPC}[c][c]{\scriptsize{}{OPC server}}
\psfrag{Controlador}[c][c]{\footnotesize{controller}}
\psfrag{Condominio_Avicola}[c][c]{\tiny{\textbf{Poultry Condominium}}}
\psfrag{Sistema_Inteligente}[c][c]{\tiny{\textbf{Intelligent System}}}
\psfrag{Alarmes}[c][c]{\tiny{alarms}}
\psfrag{Start}[c][c]{\tiny{Start}}
\psfrag{Historico}[c][c]{\tiny{history}}
\psfrag{Online_monitor}[c][c]{\tiny{online monitoring}}
    \centering
	\includegraphics[width=\linewidth]{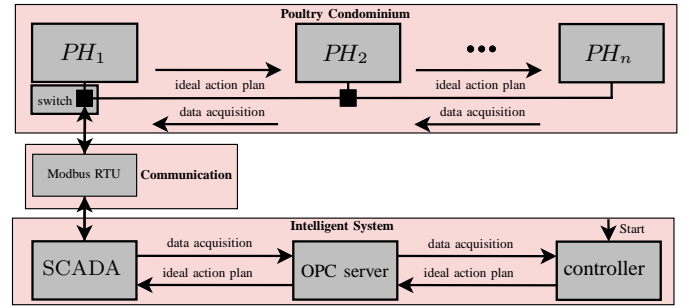}
	\caption{Layout of the proposed control architecture and data communication framework for poultry condominiums.}
\label{fig:proposta}
\end{figure}

In order for poultry houses to exchange information, adapt, and cooperate with each other, it is essential for them to be exposed as networked and communicating agents. 
In this section, a communicating \PC is discussed in the context of the methodology shown in \Fig~\ref{fig:proposta}. 

The methodology guides the integration of the process agents and results in a system that provides daily customisation for the entire production, based on the most successful case of poultry management withing a group. 

Our framework is composed of a \emph{controller}, an \emph{OPC server}, a \emph{\SCADA} application, and a \emph{communication network}.
The process starts with the controller calculating the ideal action plan based on the Algorithm~\ref{alg:alggenetic_adaptive} and in the \LSTM Network of Algorithm~\ref{alg:LSTM_implementation}, which is trained with all the dataset stored on the controller. \painter{
The controller sends this action plan to an OPC server \cite{mahnke2009opc}, which stores and accesses a \SCADA framework whenever necessary and has the purpose of facilitating the interface between SCADA and controller. A \emph{Matlab}-compliant implementation of the communication between OPC server and \SCADA application can be found in \cite{Imp_2_2020}.} 


{\painterX
Once a day, the \SCADA transmits the ideal action plan to all poultry houses through a communication network. 
In the opposite direction, but over the same infrastructure, the environment conditions (\painter{temperature, humidity, weight, feed consumption}) are captured from all poultry houses through a data acquisition system, and reported to the \SCADA system. Consequently, they become observable, via OPC server, to the controller that updates the database.


As usual control technologies for \houses are quite limited to store details about a flock, we program our own, open-sourced, software solution that works integrated with the controller. It observed data acquisition transmitted through the \SCADA system (see \Fig~\ref{fig:proposta}), customises whatever expected data and format the user wants, and stores it into a relational database, separating flocks and related attributes. 
If the sensors network is to be further expanded inside a house, to capture additional information, the database can be quickly and straightforwardly extended to receive the new data collection. 
}

\subsection{Communication infrastructure}\label{subsec:comm}

{\painterX 
The network infrastructure proposed in this paper is based on the \emph{Modbus} protocol, running in \emph{Remote Terminal Unit} (RTU) transmission mode, over a \emph{Ethernet} architecture standard in the physical layer \cite{wilamowski2018industrial}. 
Modbus uses a \emph{master-slave} model for communication. In our application, the \SCADA system is configured as the master, and each \house is a slave. In this way, data packages are transmitted, both ways, between \SCADA and \houses through the physical layer Ethernet, following the Modbus RTU protocol.
 
Each \house has a \emph{switch}, which allows multiple branches of the network to be split at several points, whenever necessary. 
This grants each \house with its own network point, marked by a distinct IP address. 
In each \house, a Ethernet - RS 232 converter receives data packages from the \emph{switch} and converts them to the serial RS - 232 standard, thus providing compatible interconnection with the controllers. 
} 

The main advantage of this configuration relies on the flexibility and simplicity in which new communication points can be quickly added or removed over the infrastructure. 
In poultry condominiums, those features can be decisive, as the number of \houses \painter{may change} frequently. 

\painter{
Due to the master-slave communication model, a slave (\PH[n]) does not initiate any type of communication in the physical environment until it has been requested to do so, by the master (\SCADA).}
For example, the \SCADA system sends a request to \PH[1], which receives and sends a reply. 
Only upon receiving a reply from \PH[1], the \SCADA can send another request, to another \PH[n]. 
This ensures that only one \house can communicate with the master at a time, which simplifies the architecture and eliminates possible transmission conflicts.

\painter{In this way, information goes} from one \house to another until its final destination, that physically characterises the network topology as a tree. \painter{This captures a poultry condominium layout quite similarly. 
In the physical infrastructure, a wired-based interconnection has been assumed as well defined due to its straightforwardness, reliability, and low cost to implement. Despite possible limitations (distance, for example), it is unlikely that a \house is more than a hundred meters apart from another, favouring a cascaded wire-based network. Furthermore, electricity power restrictions are quite unusual in \houses.}

\subsection{Supervision system}\label{subsec:super}

{\painterX \SCADA~applications allow processes to be monitored, tracked, recorded and interfered when convenient. Are usually part of a \SCADA system graphical interfaces, alarms, databases, reports, etc. \cite{boyer2009scada}.

In this paper, the proposed \SCADA system was developed using the software \emph{Elipse E3} \cite{elipsee3:Online}} in order to provide the specialist with online monitoring of the entire poultry condominium. 
This eliminates the need for non-standard manual notes and spreadsheets, as it concentrates all information into a single point, centralised, automatically-handled, database. 
\painter{For example, daily mortality is now informed directly to the \SCADA interface, from which it becomes available for multiple purposes.}

The \SCADA synoptic screens aim to record both daily history, and the history of already completed flocks. They also detect data abnormalities, reporting them through alarm indicators, messages or warnings. Graphically, the system shows the status of the process, which is updated online.

\begin{figure}[!htb]
\psfrag{Temperatura}[c][c]{\scriptsize{Temperature (\graus)}}
\psfrag{Umidade}[c][c]{\scriptsize{Humidity (\%)}}
\psfrag{Maximum}[c][c]{\tiny{Maximum}}
\psfrag{Average}[c][c]{\tiny{Average}}
\psfrag{Minimum}[c][c]{\tiny{Minimum}}
\psfrag{Historico}[c][c]{\scriptsize{History}}
\psfrag{Alarmes}[c][c]{\scriptsize{Alarms}}
\psfrag{Action_Plan}[c][c]{\scriptsize{Action Plan}}
\psfrag{Dia_lote}[c][c]{\scriptsize{Flock Day}}
\psfrag{Mortalidade}[c][c]{\scriptsize{Mortality}}
\psfrag{30}[c][c]{\tiny{30.3}}
\psfrag{28}[c][c]{\tiny{28.5}}
\psfrag{31}[c][c]{\tiny{31.7}}
\psfrag{40}[c][c]{\tiny{40}}
\psfrag{50}[c][c]{\tiny{50}}
\psfrag{60}[c][c]{\tiny{60}}
\psfrag{33}[c][c]{\tiny{30.8}}
\psfrag{52}[c][c]{\tiny{52}}
\psfrag{6}[c][c]{\tiny{6}}
    \centering
	\includegraphics[width=\linewidth]{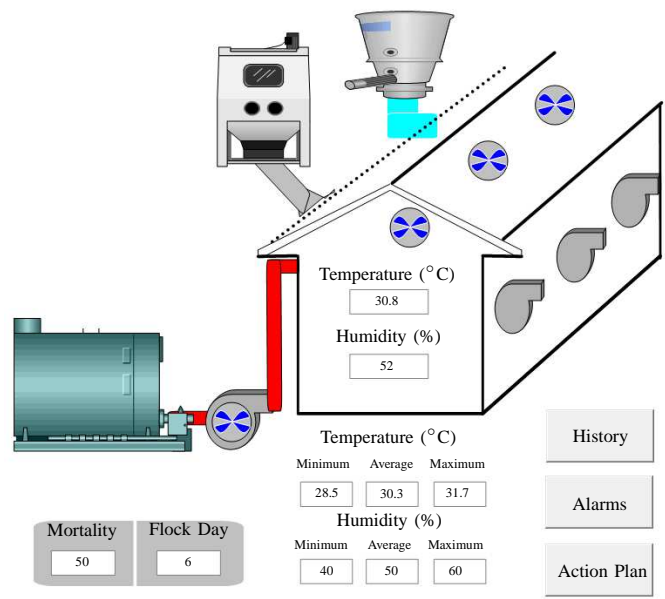}
	\caption{Layout of the implemented \SCADA system.}
\label{fig:scada_ilustration}
\end{figure}

Among the displayed information, we highlight temperature and humidity of each \house belonging to the condominium. 
Among the functionalities, we mention the possibility for the specialist to observe the action plan that has been calculated and applied at every step throughout a flock production. 
\Fig~\ref{fig:scada_ilustration} shows a condensed version of the \SCADA interface application developed in this paper. 
\painter{More complete material can be found in \cite{Imp2020}.} 






\subsection{Adaptive action plans}\label{subsec:adap}

We initially use 12 samples, corresponding to 12 produced flocks, as our real dataset, which has been split for training and testing the \LSTM models. 

{\painterX 
Here, however, we claim that this dataset can be updated in such a way that it adds customised information and become adaptive. 
Hopefully, such updates embody knowledge about specific events and features of a given \house or, equivalently, \PC. 
Then, by retraining the \LSTM models and applying the Algorithm \ref{alg:alggenetic_adaptive}, one can personalise a new strategy for managing each different production scenario. 
This approach 
makes our proposal sensitive to genetic changes, sexing, reformulation of feed recipe, new breeds, heterogeneous climate changes, etc.

The proposed adaptation is based on the following method:
\begin{enumerate}
\item \label{st:i}
Perform production using the action plans estimated by our model upon the initial dataset.

\item \label{st:ii}
By the end of a flock, collect the resulting sample. 

\item \label{st:iii}
Process the sample in order to make sure it is valid. 
It may happen that a flock diverges substantially from others, as a result of \emph{outliers} arising from atypical events, such as mortality peaks caused by sound outbreak.  
In this case, the whole sample should be deleted. Otherwise, join it with the dataset in step \ref{st:i}.

\item  \label{st:iv}
Repeat steps \ref{st:ii} and \ref{st:iii} until updating at least 25\% of the dataset in step \ref{st:i}. 
Less than 25\% is considered insufficient to capture particular features of \PCs.  
As in this paper we use 12 initial samples, we need to observe at least 3 more flocks for the dataset start to customise.   
\item \label{st:v}
Now, the need for the \LSTM readaptation is assessed. 
Use the recently recorded samples (the 3 
extra flocks) as input to the 6 \LSTM models (as in \Fig~\ref{fig:arquitetura_rede}). 
Compare the output of the \LSTM with the real values that have been stored in the database. 
A significant difference between them suggests the need for retraining of the \LSTM, and consequent reintegration with the \GA in Algorithm \ref{alg:alggenetic_adaptive}. 
Otherwise, the \LSTM is accurate for the updated dataset. 


\end{enumerate}

In conjunction, periodic dataset updates, systematic execution of the proposed bio-inspired method, and availability of adequate infrastructure resources, allow action plans to be automatically set up to the controller, straightforwardly followed by the farmer, and intelligently evolve over the time. 


We highlight that, in this paper, we do not concentrate on detecting and removing outliers from samples, as this is beyond our scope. 
We only detect them roughly, by crossing different samples, from different flocks, before they are effectively used for \LSTM training.   
More sophisticated methods, such as in \cite{yang2017robust}, can complement our approach with more detailed techniques to detect and react to possible outliers, filtering their impacts on network learning. 
}

\subsection{Assumptions and limitations}

{\painterX
The applicability of the proposed method is subject to the following premises: 
(i) we considered a period of 40 days for the flock. This can be different, as long as the \LSTM models are adjusted accordingly. 
(ii) the tested flocks included only male broilers, so that tests should be repeated for female; 
(iii) we tested only \emph{Ross}-breed broilers. We believe that our approach is applicable over any other breed, as long as the parameters (such as \FCR[\opt]) are altered accordingly; 
(iv) A \PC may include one, or any finite number of \houses, subject to physical infrastructure limitations (such as network communication), without affecting the convergence of our method;
(v) broilers are considered to arrive at the same time, under the same conditions (age, for example), for all \houses to be evaluated. 
Different setups are possible, as long as the \SCADA system is customised to consider distinct action plans concurrently. 
This is rather simple to be implemented in \SCADA by using the notion of \emph{recipe} to be automatically triggered. 
(vi) it has been also assumed that the control system is always capable of stabilising temperature and humidity as recommended by our method. This implies in orchestrating the precise actuation of exhausters, heaters, humidifiers, and other components. Details can be found in \cite{lorencena2019framework}. 


Remark that hardware and software used in this paper are both low-cost options. 
We focused on open source software, including the \SCADA, which however may not be open sourced in more complete and advanced releases. 
In terms of hardware, the cost amounts from a server, a microcontroller, a variable number of switches and other minor expenses to build the network infrastructure. In comparison to the cost involved in building an entire poultry house, implementing this innovation proposal can be considered a minor addendum.
}



\section{Conclusion}\label{sec:conc}

This paper presents an approach that combines deep learning and genetic algorithms to estimate adaptive action plans for poultry farming. We exploit a model of poultry farming, named condominium, that interconnects multiple poultry houses and allows us to both, collect distributed data, and propagate optimised action plans throughout a network infrastructure, commanded by a central supervision system. 

It has been shown that the perception of an \painter{specialist}, about some key poultry process decisions, can be captured and reproduced, to some extent, by using our bio-inspired method, and this increases productivity. A software solution implementing our method is also provided and it applies to any number of interconnected poultry houses, including one, when it fits as a particular case. 

Ongoing research focuses on technology transfer and partnership with companies for practical tests in real scale. 
We also intend to expand the set of events collected from the poultry business, which may allow to support a more complete decision making process. 
\painter{This idea aligns with the capture of climatic events outside a poultry house and to the possibility of involving \emph{cost} and \emph{profit} estimations, attempting to stratify the impact of management choices, similarly to what it has already been addressed in the swine industry \cite{fernandez2020energy}. Energy saving \cite{baierusing} is another variable to be considered in future extensions}. 




\bibliographystyle{IEEEtran}
\bibliography{ref}

\end{document}